\documentclass{article}

\usepackage[main, final]{neurips_2026}

\usepackage[utf8]{inputenc}
\usepackage[T1]{fontenc}
\usepackage{url}
\usepackage{booktabs}
\usepackage{amsfonts}
\usepackage{nicefrac}
\usepackage{microtype}
\usepackage[dvipsnames]{xcolor}
\usepackage{xspace}
\usepackage{multirow}
\usepackage{amssymb}
\usepackage{amsmath}
\usepackage{graphicx}
\usepackage{algorithm}
\usepackage{algpseudocode}
\usepackage{bm}

\definecolor{citecolor}{HTML}{0071BC}

\usepackage[breaklinks=true,colorlinks,bookmarks=false,citecolor=citecolor,linkcolor=blue,urlcolor=gray]{hyperref}

\usepackage{cleveref}

\algrenewcommand\algorithmicrequire{\textbf{Input:}}
\algrenewcommand\algorithmicensure{\textbf{Output:}}

\newcommand{\name}{\textsc{Found in Conversation}\xspace}
\newcommand{\ours}{\textsc{FiC}\xspace}
\newcommand{\method}{View-Asymmetric Self-Distillation\xspace}
\newcommand{\abbr}{VASD\xspace}

\newcommand{\lic}{Lost-in-Conversation\xspace}

\newcommand{\full}{\textsc{full}\xspace}
\newcommand{\concat}{\textsc{concat}\xspace}
\newcommand{\sharded}{\textsc{sharded}\xspace}

\newcommand{\llamasmall}{\textsc{Llama-3.2-3B}\xspace}
\newcommand{\llama}{\textsc{Llama-3.1-8B}\xspace}
\newcommand{\qwen}{\textsc{Qwen2.5-7B}\xspace}
\newcommand{\olmo}{\textsc{OLMo-2-13B}\xspace}
\newcommand{\phimodel}{\textsc{Phi-4-14B}\xspace}

\newcommand{\llamasmallid}{\texttt{Llama-3.2-3B-Instruct}\xspace}
\newcommand{\llamaid}{\texttt{Llama-3.1-8B-Instruct}\xspace}
\newcommand{\qwenid}{\texttt{Qwen2.5-7B-Instruct}\xspace}
\newcommand{\olmoid}{\texttt{OLMo-2-1124-13B-Instruct}\xspace}
\newcommand{\phiid}{\texttt{phi-4}\xspace}

\newcommand{\promptrule}{\textsc{prompt-rule}\xspace}
\newcommand{\promptselfcheck}{\textsc{prompt-selfcheck}\xspace}
\newcommand{\gatedself}{\textsc{gated-self}\xspace}
\newcommand{\gatedexternal}{\textsc{gated-external}\xspace}

\title{\name: LLMs Teach Themselves to Close the Multi-Turn Gap}

\author{%
  Tianlang Chen \\
  Stanford University \\
  \texttt{tlchen@cs.stanford.edu} \\
  \And
  Shirley Wu \\
  Stanford University \\
  \texttt{shirwu@cs.stanford.edu} \\
  \And
  Jure Leskovec \\
  Stanford University \\
  \texttt{jure@cs.stanford.edu} \\
}

\begin{document}

\maketitle

\begin{abstract}
\looseness=-1
Large Language Model (LLM) interactions are typically underspecified, with users clarifying all necessary details across multiple conversational turns.
Yet recent work shows that LLMs perform far worse in this multi-turn setting than in a single turn with the same information available at once, a phenomenon termed ``\lic.''
However, bridging this gap effectively remains an open problem.
Here we introduce \name (\ours), a training framework where a model teaches itself to find and recover its single-turn competence given underspecified multi-turn prompts.
We develop \method, which distills across two views of the same task information—a single-turn view for the teacher, a multi-turn view for the student—transferring strong single-turn behavior into weak multi-turn behavior. This requires no stronger external teacher, which is unavailable as even frontier LLMs exhibit this gap.
Across model families (Llama, Qwen, Phi, and OLMo) and sizes (3B–14B), \ours attains 92--100\% of single-turn performance, yielding more efficient and helpful multi-turn conversations with single-turn capabilities largely intact.
\end{abstract}

\section{Introduction}
\label{intro}

\looseness=-1
Large Language Models (LLMs) reliably produce high-quality responses when given a fully specified, single-turn prompt~\citep{GPT-4, Claude, llama3,yang2025qwen3}.
However, real users seldom interact this way. They typically start with an underspecified request and provide further details to clarify their needs across subsequent turns, which is a common feature of human communication~\citep{zipf1949human,ferreira2008ambiguity,Frisson2009SemanticUI} and the dominant mode of real-world LLM use~\citep{d2022underspecification,herlihy2024overcoming}.
Yet recent work shows that LLMs handle this setting strikingly poorly: across fifteen leading open- and closed-weight models, multi-turn underspecified conversations incur an average 39\% drop in task performance compared to single-turn prompts carrying the same information, a phenomenon termed \emph{\lic}~\citep{laban2025llms}.

\looseness=-1
Closing this multi-turn gap proves difficult, since each natural class of methods runs into distinct obstacles.
Inference-time methods such as prompting often fail to overturn ingrained model behavior~\citep{huang2023large,sharma2024towards}, are brittle to prompt design~\citep{sclar2024quantifying,verma2024brittle}, and incur latency that hinders real-time deployments~\citep{mu2023learning,wu2025inference}.
Post-training methods face their own difficulties: Supervised fine-tuning (SFT) alone is prone to distribution shifts and shortcut learning, with models merely memorizing surface patterns rather than acquiring generalizable behavior~\citep{geirhos2020shortcut,chu2025sft,lampinen2025generalization}.
Reinforcement learning (RL) in this multi-turn setting typically suffers from intractable credit assignment over long trajectories due to sparse reward signals~\citep{abdulhai2025lmrl,lu2025onpolicydistillation,hubotter2026reinforcement}, and forward-sampling costs that scale with conversation length~\citep{wu2025collabllm,lu2025scaling}.
Distillation provides a dense, token-level learning signal that sidesteps RL's sparse-reward problem, but typically requires a stronger teacher~\citep{guo2025deepseek,yang2025qwen3,lu2025onpolicydistillation}, yet even frontier models struggle in this setting~\citep{laban2025llms}.
Closing this multi-turn gap therefore calls for a new training framework.

\begin{figure}[t]
\centering
\includegraphics[width=\textwidth]{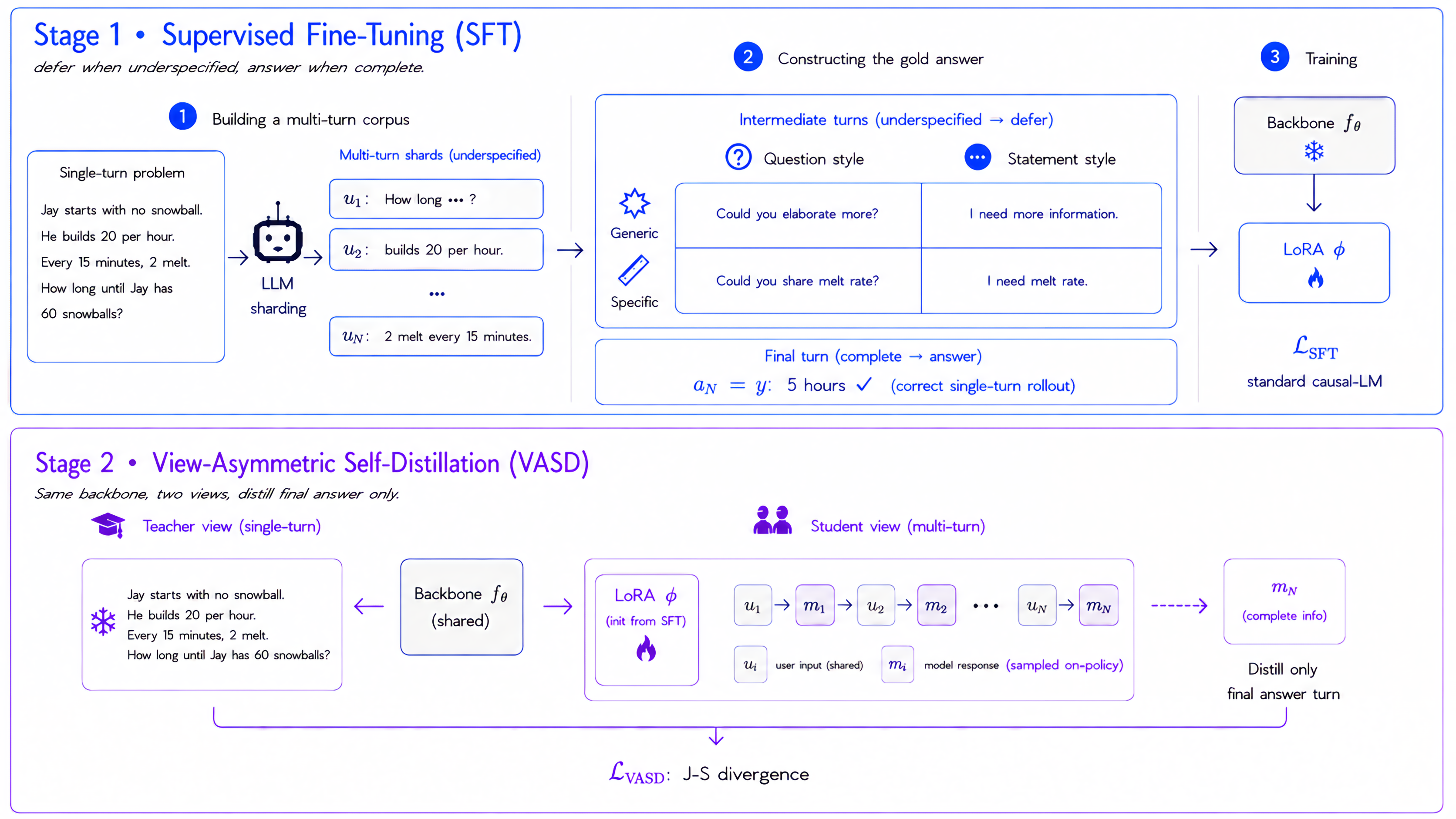}
\caption{\looseness=-1 Overview of \name. SFT warm-starts the model: build a multi-turn corpus from a standard single-turn benchmark, construct per-turn gold answers with diversity, and train with a standard causal-LM loss. \abbr runs the same backbone on two information-equivalent views: a teacher view on the single-turn setting and a student view on the multi-turn setting with intermediate responses sampled on-policy. A token-level distillation loss aligns the student to the teacher.}
\label{fig:method}
\end{figure}

\looseness=-1
Here we introduce \name (\ours), a general training framework that enables a model to find its way back to its native single-turn ability in multi-turn conversations (\Cref{fig:method}).
Our motivating observation is that the same model performs better given identical information in a single turn. This indicates the gap is behavioral rather than capability-bound: the competence is present inside the model but struggles to surface in multi-turn conversations.
\ours recovers this latent competence in two complementary stages.
First, SFT on a carefully curated multi-turn corpus trains the model to defer or request clarification when information is missing, and to commit to an answer only once all necessary details are given.
In the second stage, we develop a novel \method (\abbr) method, which uses the model's strong single-turn behavior to teach its own weaker multi-turn behavior.
\abbr runs the same backbone on two views of the same problem: a teacher pass on the single-turn view, and a student pass on the multi-turn view.
A token-level distillation loss on the final answer, where both teacher and student have complete information, aligns the student's distribution with the teacher's.

\looseness=-1
\abbr is the core technical contribution of \ours.
By inheriting distillation's dense, token-level signal, \abbr circumvents the sparse-reward credit assignment problem that typically constrains multi-turn RL~\citep{abdulhai2025lmrl,lu2025onpolicydistillation}. By adopting a self-distillation form, \abbr bypasses standard distillation's requirement of a stronger external teacher, which is unavailable in this setting.
\abbr exploits a novel asymmetry between teacher and student.
Prior self-distillation methods generally use an \emph{information asymmetry} as the learning signal: the teacher is conditioned on privileged information such as external feedback~\citep{hubotter2026reinforcement,song2026expanding} or expert demonstrations~\citep{zhao2026self,shenfeld2026self} the student does not see. However, such privileged information is not available in our setting.
\abbr instead introduces a novel \emph{input-view asymmetry}: teacher and student share the same backbone and task information, differing only in how that information is presented. This yields a clean learning signal, since the teacher–student gap is exactly the multi-turn-vs-single-turn gap we want to close, with no other distributional difference.
In sum, \abbr bridges the multi-turn gap internally via dense self-distillation signals without requiring stronger external teachers.

\looseness=-1
We evaluate \ours on the multi-turn benchmark of~\citet{laban2025llms}\ across diverse model families (Llama~\citep{llama3}, Qwen~\citep{qwen2.5}, Phi~\citep{phi4}, and OLMo~\citep{olmo2}) and sizes (3B–14B).
On the multi-turn Math task, \ours attains $92$--$100\%$ of single-turn performance, closing $75$--$100\%$ of the multi-turn gap, while leaving single-turn capabilities largely intact.
\ours also enables more efficient conversations with up to $33\%$ fewer tokens.
The trained checkpoints further transfer to Database and Actions tasks with minimal adaptation (at most $6$ A100-hours), recovering most of the gap.
Together, the results demonstrate that the multi-turn gap is largely closeable from inside the model with our \ours framework.

\begin{figure}[t]
\centering
\includegraphics[width=.85\textwidth]{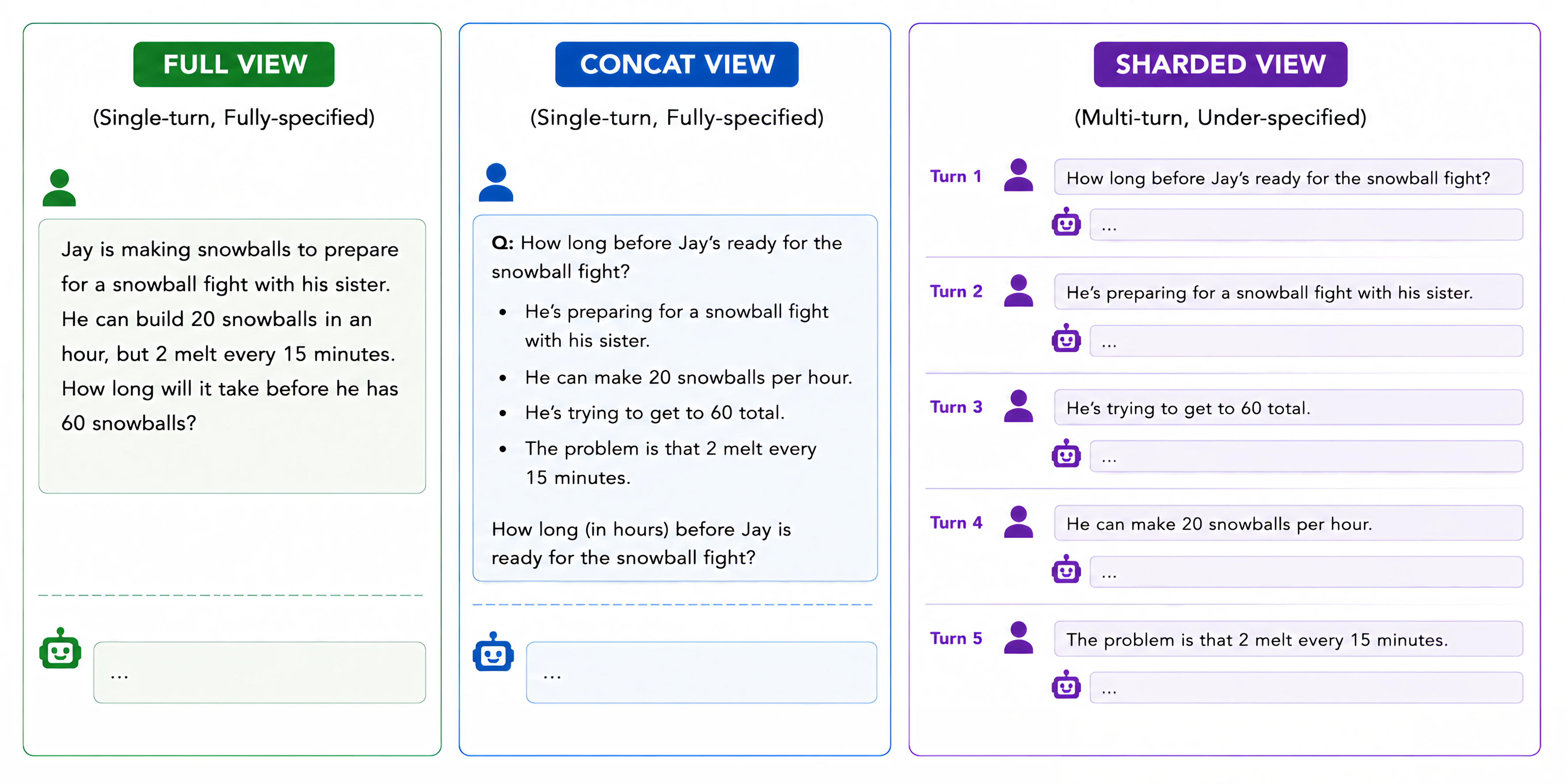}
\caption{Three information-equivalent views of the same instruction. \full (left) presents the original single-turn instruction from a standard benchmark. \concat (middle) preserves the single-turn structure but replaces the prose with a bullet list of shards. \sharded (right) reveals one shard per turn in a multi-turn conversation — the underspecified setting where \lic occurs.}
\label{fig:view}
\end{figure}

\section{Problem Formulation}
\label{sec:formulation}

\looseness=-1
A multi-turn conversation can be modeled as a sequence $\{t_i\}_{i=1}^{N}$ of $N$ turns, where each turn $t_i := (u_i, m_i)$ pairs a user input $u_i$ with a model response $m_i \sim \pi\!\left(\,\cdot \mid t_{1:i-1} \cup \{u_i\}\right)$. To isolate and quantify the difficulty that multi-turn delivery itself introduces, we consider conversations whose user inputs $\{u_i\}_{i=1}^{N}$ admit an \emph{information-equivalent single-turn counterpart}. Specifically, let $I(\cdot)$ be the task-relevant information of a prompt. A single-turn prompt $q$ is information-equivalent to multi-turn $\{u_i\}_{i=1}^{N}$ when $I(q) = \bigcup_{i=1}^{N} I(u_i)$. We further focus on the \emph{underspecified} regime, where $I(u_i) \subsetneq I(q)$ for every $i$ --- no single user turn is sufficient to complete the task. Under this setup, the model's performance on single-turn $q$ and on multi-turn $\{u_i\}_{i=1}^{N}$ is directly comparable: any gap reflects how the same information is delivered, not what is delivered. Let $\bar{P}(\pi; \cdot)$ denote the model's expected task score on an input set. \citet{laban2025llms} showed that for an information-equivalent pair $(q, \{u_i\}_{i=1}^{N})$, $\bar{P}(\pi; \{u_i\}_{i=1}^{N}) \ll \bar{P}(\pi; q)$ holds across diverse open-source and frontier LLMs (\lic). This setting is distinct from episodic multi-turn paradigms such as chain-of-thought reasoning, where the task is fully specified from the start, and the multiple turns are just intermediate reasoning steps that combine into the final answer.

\looseness=-1
Given a model $\pi$, our goal is a training framework $T : \pi \mapsto \pi'$ that closes the gap, $\bar{P}(\pi'; \{u_i\}_{i=1}^{N}) \to \bar{P}(\pi; q)$, while preserving single-turn performance, $\bar{P}(\pi'; q) \approx \bar{P}(\pi; q)$.
We define the \emph{recovery rate} $R := \big(\bar{P}(\pi'; \{u_i\}_{i=1}^{N}) - \bar{P}(\pi; \{u_i\}_{i=1}^{N})\big) \,/\, \big(\bar{P}(\pi; q) - \bar{P}(\pi; \{u_i\}_{i=1}^{N})\big)$, the fraction of the original multi-turn gap that training closes, and the \emph{single-turn attainment} $A := \bar{P}(\pi'; \{u_i\}_{i=1}^{N}) \,/\, \bar{P}(\pi; q)$, the fraction of the original model’s single-turn performance that the trained model attains under multi-turn delivery. We restrict to training procedures without supervision from stronger external models. This restriction is methodological, not fundamental: it makes single-turn performance $\bar{P}(\pi; q)$ a principled ceiling (so $R \le 1$ and $A \le 1$) and thereby isolates how much of the multi-turn gap is closed by the training framework itself, separately from any capability injected by an external teacher. In practice, the framework can freely incorporate stronger models as supervision sources to further enhance the performance.

\pagebreak

\section{\name Framework}
\label{sec:method}

\looseness=-1
\name (\ours) is a general framework where the model teaches itself to recover its single-turn
competence in multi-turn conversation. The first stage of training is supervised fine-tuning (SFT) on a multi-turn corpus that we carefully construct. This stage acts as a behavioral warm-start: it teaches the model to refuse or ask for clarification when information is incomplete (\Cref{sec:method:sft}). For the second stage we develop a novel \method~(\abbr), which runs the same backbone on two information-equivalent views
of the same problem  (\Cref{sec:method:vasd}). \abbr aligns the student's multi-turn-view distribution
with the teacher's single-turn-view distribution at the token level with distillation. The two stages are complementary: SFT makes the model a competent partial respondent, whereas \abbr is where the multi-turn-vs-single-turn gap is actually closed.

\subsection{SFT for Grounded Context}
\label{sec:method:sft}

\looseness=-1
\lic is often driven by premature answer commitment~\citep{laban2025llms}: models fill missing conditions with hallucinated assumptions and then anchor on them in subsequent turns (\Cref{fig:qualitative} left), a behavior reinforced by standard post-training that optimizes for immediate next-turn rewards~\citep{wu2025collabllm}.
Besides producing unhelpful answers and wasting tokens, this also corrupts the conditioning context that the model carries into the rest of the conversation.
The first stage of \ours therefore targets this behavior directly: the model is trained with standard SFT on a curated multi-turn corpus to defer incomplete queries, ask for specific missing details when they are predictable, and commit to an answer only once enough information is provided.
SFT serves as a behavioral primer that produces conversations with grounded context for the \abbr stage to operate on.

\looseness=-1
\paragraph{Building a multi-turn corpus.}
We follow the recipe of~\citet{laban2025llms} to construct a multi-turn corpus from a standard single-turn benchmark. Given a fully specified single-turn instruction $q$, a sharding procedure prompts an LLM to decompose $q$ into a set of shards $\{u_i\}_{i=1}^{N}$ that is jointly information-equivalent to $q$, with each individual shard strictly less informative (\Cref{sec:formulation}).
To verify information-equivalence empirically, we form a single-turn prompt $q_{\text{concat}} := \mathrm{concat}(u_1,\ldots, u_N)$ that lists all shards as bullet points, and accept the sharding when the model's performance on $q_{\text{concat}}$ matches its performance on $q$, both in the single-turn setting.
We refer to these three presentations of the same underlying problem as the \textbf{\full} view ($q$, the original single-turn instruction), the \textbf{\concat} view ($q_{\text{concat}}$,
single-turn with concatenated shards), and the \textbf{\sharded} view ($\{u_i\}_{i=1}^{N}$, multi-turn); all three are information-equivalent by construction (\Cref{fig:view}).
Since \citet{laban2025llms} released only \emph{test} shards, we apply their procedure to a disjoint pool to produce \emph{training} shards for our experiments.

\paragraph{Constructing the gold answer.}
\looseness=-1
A training example is a chat-template-formatted trace
$(\mathrm{system},\, u_1,\, a_1,\, u_2,\, a_2,\, \ldots,\, u_N,\, a_N)$,
where the gold assistant answer $a_t$ must behave one way at intermediate turns when information is incomplete, and another at the final turn when it is complete.

\looseness=-1
\emph{Intermediate turns.} $a_t$ ($t < N$) should act as a deferral rather than an answer attempt. A single fixed wording of $a_t$ reused across turns induces the surface-pattern collapse SFT is most prone to~\citep{chu2025sft}, so we diversify $a_t$ along two axes. The first is \emph{stylistic diversity}: across turns and examples, $a_t$ varies in format, phrasing, register, and length, ensuring the model learns the underlying deferral behavior rather than memorizing a specific phrase. The second is \emph{position-aware specificity}. In early turns, when context is too sparse for the model to plausibly identify what is missing, $a_t$ is a generic deferral. In later turns, as the context becomes clearer, $a_t$ shifts to explicitly requesting the specific missing information, which is more helpful to users. To ensure honest grounding, a candidate $a_t$ is admitted only if a judge confirms its requested information is both missing and necessary. The entire pipeline above is automated by prompting LLMs as tooling, with negligible human effort: following \citet{laban2025llms}, GPT-4o shards the instructions, and GPT-4o-mini phrases and verifies the deferrals, which carry no task content since unrevealed values are abstracted away. External LLMs thus enter \ours only as corpus-construction tooling in this stage and never write a final-answer target.
Together, these two axes shape a corpus that supports learning grounded, helpful behavior at intermediate turns.

\looseness=-1
\emph{Final turn.} $a_N$ is drawn from the model's own correct rollouts of the same problem under the single-turn \concat view.
This keeps the SFT signal close to the model's prior distribution, avoiding the distribution shift that would arise from supervising on a generic reference answer.
Additionally, the presence of multiple correct \concat rollouts per problem yields natural final-turn diversity: a different correct rollout is supplied each SFT epoch, preventing collapse onto a single canonical phrasing.

\begin{algorithm}[t]
\caption{View-Asymmetric Self-Distillation (\abbr)}
\label{alg:vasd}
\begin{algorithmic}[1]
\Require Frozen backbone $\pi_\theta$; LoRA adapter $\phi$ initialized from SFT checkpoint; multi-turn corpus $\mathcal{D} = \{(\{u_i\}_{i=1}^{N},\, q_{\text{concat}})\}$; learning rate $\eta$.

\For{each training step}
    \State Sample $(\{u_i\}_{i=1}^{N},\, q_{\text{concat}}) \sim \mathcal{D}$
    \For{$i = 1, \ldots, N-1$}
        \State $m_i \sim \pi_{\theta+\phi}\!\left(\,\cdot \mid t_{1:i-1} \cup \{u_i\}\right)$ \Comment{$t_j := (u_j, m_j)$, gradients off}
    \EndFor
    \State $y \sim \pi_\theta(\,\cdot \mid q_{\text{concat}})$ \Comment{or precomputed; gradients off}
    \State $p^{(s)}_t \gets \pi_{\theta+\phi}(T_s)$ at positions $t \in \mathcal{Y}$ \Comment{$T_s := (\mathrm{sys}, u_1, m_1, \ldots, u_N, y)$; gradients on}
    \State $p^{(t)}_t \gets \pi_\theta(T_t)$ at positions $t \in \mathcal{Y}$ \Comment{$T_t := (\mathrm{sys}, q_{\text{concat}}, y)$; gradients off}
    \State $\mathcal{L}_{\abbr} \gets \dfrac{1}{|\mathcal{Y}|} \displaystyle\sum_{t \in \mathcal{Y}} D_{\mathrm{JS}}\!\left(p^{(t)}_t \,\big\|\, p^{(s)}_t\right)$
    \State $\phi \gets \phi - \eta \nabla_\phi \mathcal{L}_{\abbr}$ \Comment{update adapter only}
\EndFor
\State \Return $\pi_{\theta+\phi}$
\end{algorithmic}
\end{algorithm}

\subsection{\method~(\abbr)}
\label{sec:method:vasd}
\looseness=-1
SFT successfully instills deferral, but the model afterward tends to over-defer, a well-documented consequence of off-policy, sequence-level supervision~\citep{agarwal2024policy,chu2025sft}.
Closing this remaining gap requires a denser learning signal.
Distillation is a natural choice, but typically requires a stronger external teacher that is unavailable here: even frontier models suffer the same multi-turn performance degradation~\citep{laban2025llms}, and a vanilla distillation from a stronger teacher fails to close the gap (\Cref{tab:ablation-vanilla}).
To this end, we propose \method (\abbr), which bypasses the need for an external teacher.
Our key insight is that a competent teacher already lives \emph{inside} the model: the same checkpoint that struggles on the multi-turn \sharded view is strong on the information-equivalent single-turn \concat view.
No external model is involved at this stage.
\abbr makes this internal expertise explicit by running the same backbone on both views in parallel, aligning the student’s distribution (on the multi-turn \sharded view) with the teacher’s (on the single-turn \concat view) at the token level.

\looseness=-1
\Cref{alg:vasd} outlines \abbr.
The student runs on the \sharded view: beginning with the system prompt and the initial user query $u_1$, the student generates its own multi-turn trajectory where each intermediate response $m_i$ is sampled on-policy from the current student conditioned on the conversation so far.
The teacher runs on the \concat view, conditioned on the system prompt and single-turn $q_{\text{concat}}$ alone.
Supervision is performed only at the final turn, where both teacher and student have complete task information. The intermediate responses  are part of the student's input so attention can condition on them, but they receive no gradient signal.
The loss is a token-level Jensen--Shannon divergence~\citep{jsdloss} between the student's and teacher's distributions over the answer span,
\begin{equation}
\label{eq:vasd-loss}
\resizebox{.93\linewidth}{!}
{
$\mathcal{L}_{\abbr}
\;=\;
\frac{1}{|\mathcal{Y}|} \sum_{t \in \mathcal{Y}}
D_{\mathrm{JS}}\!\left(p^{(t)}_t \,\big\|\, p^{(s)}_t\right),
\quad
D_{\mathrm{JS}}(p \,\|\, q)
\;=\;
\tfrac{1}{2} D_{\mathrm{KL}}(p \,\|\, m) + \tfrac{1}{2} D_{\mathrm{KL}}(q \,\|\, m),
\;\; m = \tfrac{1}{2}(p + q),$
}
\end{equation}
where $\mathcal{Y}$ indexes the answer-span positions and
$p^{(s)}_t, p^{(t)}_t$ are the student's and teacher's next-token distributions at position $t$.

\looseness=-1
Two design properties jointly make \abbr effective.
First, \emph{an aligned learning signal.} Teacher and student share the same backbone and, since the two views are information-equivalent by construction, the same task information, so the teacher--student gap reduces to exactly the multi-turn-vs-single-turn gap we want to close, with no capacity gap to bridge and no distributional difference for the student to learn around.
Second, \emph{intermediate responses are sampled on-policy.} The student learns to commit to a correct answer after its own training-time intermediate generations rather than static off-policy traces (cf.~\Cref{tab:ablation-design}).
Together, these properties yield a dense, token-level supervision signal aligned exactly with the gap of interest.

\pagebreak

\section{Experiments}
\label{exp}

\subsection{Experimental Setup}
\label{sec:exp:setup}

\paragraph{Benchmark.}
\looseness=-1
We evaluate \ours on the sharded multi-turn benchmark of~\citet{laban2025llms} across different domains.
\textbf{Math} draws from GSM8K~\citep{cobbe2021gsm8k}, where the assistant must solve a grade-school word problem. \ours checkpoints are trained on the math domain.
To test cross-domain generalization, we also evaluate on two \emph{out-of-domain} tasks: \textbf{Database}, sourced from Spider~\citep{yu2018spider}, where the assistant writes an SQL query given a database schema and a natural-language request; and \textbf{Actions}, drawn from the Berkeley Function Calling Leaderboard~\citep{2024bfcl}, where the assistant emits programmatic API calls given a tool specification and a user instruction.
Each original single-turn instruction $q$ is paired with an information-equivalent set of shards $\{u_i\}_{i=1}^{N}$.

\paragraph{Evaluation.}
We follow the simulation protocol
of~\citet{laban2025llms}, where an LLM user simulator delivers each shard $u_i$ turn by turn. Mirroring real user behavior, it dynamically selects the next shard based on the conversation history and the assistant's previous responses.
A strategy classifier identifies and scores answer attempts against a reference. If an attempt is incorrect, the simulator continues revealing shards until a correct answer is produced or the shard set is exhausted.
Performance is evaluated under the three views (\Cref{fig:view}): \full (the original single-turn instruction), \concat (a single-turn instruction listing all shards), and \sharded (multi-turn, underspecified).
Consistent with~\citet{laban2025llms}, we run 10 simulations per instruction and report mean accuracy, with 95\% bootstrap confidence intervals in \Cref{app:ci}.
We take the single-turn \concat accuracy of the original model as the performance ceiling, since \concat and \sharded carry identical information. The \emph{recovery rate} $R$ reports how fully \ours recovers the performance lost under multi-turn delivery, and the \emph{single-turn attainment} $A$ reports how close \ours's multi-turn accuracy comes to that ceiling (cf.~\Cref{sec:formulation}).
See \Cref{app:exp-details} for more details.

\paragraph{Models.}
We evaluate \ours across five publicly released instruction-tuned models in the $3$B--$14$B parameter range and spanning four model families:
\llamasmallid and \llamaid~\citep{llama3}, \qwenid~\citep{qwen2.5}, \olmoid~\citep{olmo2}, and \phiid~\citep{phi4}.

\paragraph{Training.}
\looseness=-1
Our training corpus is constructed by applying the sharding pipeline of~\citet{laban2025llms} to a pool of GSM8K instructions disjoint from the provided test set, following the procedure described in~\Cref{sec:method:sft}.
For each model, we train a LoRA adapter~\citep{lora} on the model's own backbone, leaving all base weights frozen. Additional details are presented in~\Cref{app:exp-details}.

\paragraph{Baselines.}
\looseness=-1
Alongside original model baselines, we compare against four inference-time methods covering the two natural training-free ways to address the multi-turn gap: instructing the model to defer until all information is present, either through a system-prompt rule or a self-assessed readiness prefix, and gating its answer attempts through an auxiliary judge, either an external LLM or a second pass of the model itself. ~\Cref{app:test-time} describes the details.

\subsection{Main Results}
\label{exp:res}
\begin{table}[t]
\centering
\footnotesize
\setlength{\tabcolsep}{4pt}
\caption{Main results on the multi-turn Math task derived from GSM8K. We perform 10 runs per instruction and report average accuracy.
The \emph{recovery rate} $R$ is the fraction of the baseline \concat-to-\sharded gap closed by \ours, and the \emph{single-turn attainment} $A$ is \ours's \sharded accuracy divided by baseline \concat accuracy.
Results of inference-time intervention baselines are presented in Table~\ref{tab:prompt-baselines}.}
\label{tab:main}
\resizebox{\linewidth}{!}{%
\begin{tabular}{llccccc}
\toprule
Setting & Method & \footnotesize\llamasmall & \footnotesize\llama & \footnotesize\qwen & \footnotesize\olmo & \footnotesize\phimodel \\
\midrule
\multirow{2}{*}{\full}
 & Baseline     & $62.5$ & $72.5$ & $84.2$ & $82.9$ & $88.3$ \\
 & \ours        & $58.3$ & $73.8$ & $83.6$ & $79.5$ & $89.8$ \\
\midrule
\multirow{2}{*}{\concat}
 & Baseline     & $60.2$ & $71.5$ & $85.0$ & $82.2$ & $89.4$ \\
 & \ours        & $52.6$ & $72.3$ & $82.6$ & $77.8$ & $88.0$ \\
\midrule
\multirow{2}{*}{\sharded}
 & Baseline     & $39.2$ & $44.7$ & $58.9$ & $54.4$ & $61.2$ \\
 & \ours        & $\bm{60.4}$ & $\bm{71.7}$ & $\bm{78.7}$ & $\bm{77.1}$ & $\bm{82.4}$ \\
\midrule
\multicolumn{2}{l}{\textbf{Relative Improvement}}
                & $54\%$ & $60\%$ & $34\%$ & $42\%$ & $35\%$ \\
\multicolumn{2}{l}{\textbf{Recovery Rate} $R$}
                & $100\%$ & $100\%$ & $76\%$ & $82\%$ & $75\%$ \\
\multicolumn{2}{l}{\textbf{Single-turn Attainment} $A$}
                & \textcolor{PineGreen}{$\bm{100\%}$} & \textcolor{PineGreen}{$\bm{100\%}$}
                & \textcolor{PineGreen}{$\bm{93\%}$} & \textcolor{PineGreen}{$\bm{94\%}$} & \textcolor{PineGreen}{$\bm{92\%}$} \\
\bottomrule
\end{tabular}%
}
\end{table}
\begin{table}[t]
\centering
\small
\setlength{\tabcolsep}{4pt}
\caption{\looseness=-1 Average accuracy on Database and Actions with 10 runs per instruction. Adaptation cost is measured in A100 GPU hours. \ours generalizes to out-of-domain tasks with minimal overhead.}
\label{tab:transfer}
\resizebox{\linewidth}{!}{%
\begin{tabular}{llccccc}
\toprule
Domain & Setting & \llamasmall & \llama & \qwen & \olmo & \phimodel \\
\midrule
\multirow{5}{*}{Database}
 & \full \ (Baseline)              & $54.5$ & $68.2$ & $81.3$ & $60.0$ & $89.9$ \\
 & \concat \ (Baseline)            & $33.2$ & $65.4$ & $80.4$ & $49.9$ & $85.0$ \\
 & \sharded \ (Baseline)           & $25.3$ & $31.4$ & $39.7$ & $31.5$ & $38.5$ \\
 & \sharded \ (\ours)              & $\bm{51.1}$ & $\bm{62.0}$ & $\bm{70.9}$ & $\bm{58.3}$ & $\bm{82.2}$ \\
 \cmidrule{2-7}
 & A100-hrs (\ours)                 & $2.80$ & $1.74$ & $3.67$ & $6.00$ & $4.84$ \\
\midrule
\multirow{5}{*}{Actions}
 & \full \ (Baseline)              & $89.1$ & $88.2$ & $95.7$ & $58.2$ & $74.5$ \\
 & \concat \ (Baseline)            & $84.1$ & $89.6$ & $95.2$ & $62.0$ & $67.3$ \\
 & \sharded \ (Baseline)           & $37.8$ & $45.0$ & $42.4$ & $27.9$ & $29.9$ \\
 & \sharded \ (\ours)              & $\bm{76.2}$ & $\bm{84.9}$ & $\bm{85.0}$ & $\bm{65.5}$ & $\bm{89.0}$ \\
  \cmidrule{2-7}
 & A100-hrs (\ours)                 & $2.55$ & $0.72$ & $0.80$ & $1.50$ & $1.42$ \\
\bottomrule
\end{tabular}%
}
\end{table}

\paragraph{\ours closes the multi-turn \lic gap.}
\looseness=-1
\Cref{tab:main} reports performance under the three views before and after \ours training.
On the \sharded view, the baselines' performance drops by at least $31\%$, reproducing the \lic phenomenon~\citep{laban2025llms}.
\ours attains $92$--$100\%$ of baseline single-turn performance and recovers $75$--$100\%$ of the multi-turn gap.
By contrast, the four inference-time interventions fall well short of \ours and show marginal or even negative improvement over the original model (\Cref{tab:prompt-baselines}), confirming that inference-time methods alone are insufficient in this setting.

\paragraph{\ours preserves single-turn capabilities.}
\looseness=-1
Closing the multi-turn gap is only useful if it does not degrade single-turn behavior.
Comparing the \full and \concat rows of~\Cref{tab:main} before and after \ours training, we find that single-turn accuracy is preserved on every backbone of $7$B parameters or more: across these four, it differs by at most $4.4$ points, within overlapping confidence intervals (\Cref{app:ci}). The sole exception is the smallest backbone, \llamasmall, whose \concat accuracy drops by $7.6$ points after training. We attribute this to a capacity--behavior trade-off at $3$B, where the backbone has less spare capacity to absorb the new multi-turn behavior, and read the result as closing the multi-turn gap at a single-turn cost rather than as a clean recovery.
Together, the multi-turn improvement on every backbone and the single-turn preservation on all but the smallest establish \ours as an effective framework for addressing the \lic issue.

\paragraph{\ours transfers to out-of-domain tasks with minimal adaptation.}
\looseness=-1
A practical framework is expected to generalize across different tasks beyond the training domains.
Starting from each model's math-distilled \ours checkpoint, we apply a lightweight final-turn-only SFT pass with $45$--$67$ examples for Actions and $96$--$201$ examples for Database, with $0.7$--$6.0$ A100-hours per model.
No \abbr pass is run on the target domain, intermediate-turn behavior is inherited from the math checkpoint, and the final-turn pass supplies only the domain's answer format (\Cref{app:transfer}).
\Cref{tab:transfer} presents the results. \ours recovers multi-turn performance on these new domains with minimal adaptation.
Notably, on Actions \olmo and \phimodel exceed their single-turn accuracy, which we attribute to the strict format requirement: the Actions checker enforces a strict AST format that both original models frequently violate, rejecting some semantically correct responses; the SFT fixes the issue.
The fact that limited adaptation suffices suggests that what \ours learns is largely a principle for converting underspecified multi-turn deliveries into
grounded final answers, not a domain-specific solution strategy.

\subsection{Additional Analyses}
\label{exp:ana}
\begin{table}[t]
\centering
\footnotesize
\setlength{\tabcolsep}{3pt}

\begin{minipage}[t]{0.40\linewidth}
\centering
\caption{Mean tokens in multi-turn \sharded conversation.}
\label{tab:efficiency}
\vspace{2pt}
\begin{tabular}{lccc}
\toprule
Model        & Baseline   & \ours  & $\Delta$ \\
\midrule
\llamasmall  & $975$  & $978$  & $+0.2\%$  \\
\llama       & $1175$ & $979$  & $-16.7\%$ \\
\qwen        & $1693$ & $1143$ & $-32.5\%$ \\
\olmo        & $1905$ & $1292$ & $-32.2\%$ \\
\phimodel    & $1683$ & $1247$ & $-25.9\%$ \\
\bottomrule
\end{tabular}
\end{minipage}%
\hfill
\begin{minipage}[t]{0.32\linewidth}
\centering
\caption{Design choice ablation on \llama \sharded.}
\label{tab:ablation-design}
\vspace{2pt}
\begin{tabular}{lc}
\toprule
Method                              & Acc.        \\
\midrule
Baseline                            & $44.7$      \\
SFT only                            & $55.6$      \\
\;+ off-policy \abbr                & $60.0$      \\
\;+ on-policy \abbr~(\ours)         & $\bm{71.7}$ \\
\bottomrule
\end{tabular}
\end{minipage}%
\hfill
\begin{minipage}[t]{0.24\linewidth}
\centering
\caption{Vanilla distillation does not work. Llama 8B $\!\to\!$ 3B.}
\label{tab:ablation-vanilla}
\vspace{2pt}
\begin{tabular}{lc}
\toprule
Method               & Acc.        \\
\midrule
Baseline             & $39.2$      \\
Vanilla distill.     & $30.7$      \\
\ours                & $\bm{60.4}$ \\
\bottomrule
\end{tabular}
\end{minipage}

\end{table}

\begin{figure}[t]
\centering
\includegraphics[width=\textwidth]{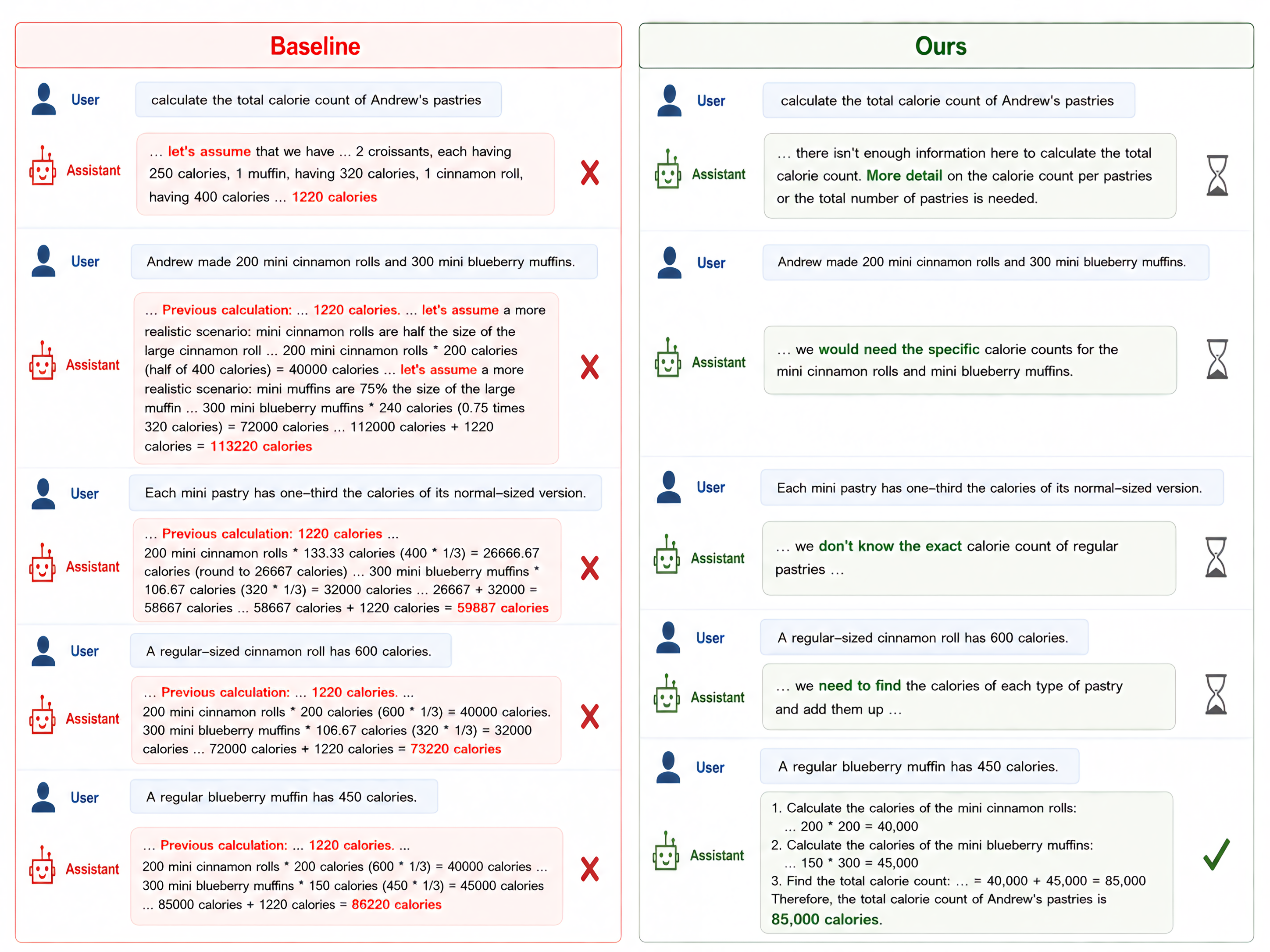}
\caption{Qualitative comparison of the \llama baseline (left) and \ours (right) on a multi-turn GSM8K problem. The baseline anchors on hallucinated assumptions in early turns and propagates them to a wrong final answer. \ours defers at intermediate turns, requesting specific missing information, and produces the correct answer once information is complete.}
\label{fig:qualitative}
\end{figure}

\paragraph{\ours produces more efficient, better-grounded conversations.}
\looseness=-1
Beyond accuracy, we analyze the efficiency and behavioral grounding. \Cref{tab:efficiency} reports mean tokens per conversation. \ours reduces token usage by up to $33\%$ while raising accuracy by up to $60\%$; \llamasmall is flat, but its baseline was already the shortest of the five. Notably, the mean number of turns is essentially unchanged ($5.27$--$5.39$ across all models and both conditions), so the savings do not come from terminating conversations early. The baseline typically responds to incomplete shards with long, ungrounded answer attempts that hallucinate the missing values, whereas \ours emits a brief deferral and waits.
\Cref{fig:qualitative} shows a representative pair of trajectories: the baseline anchors on hallucinated assumptions from earlier turns and propagates them through to a wrong final answer, while \ours actively asks for the missing information at intermediate turns and waits until all information is revealed to produce a correct answer.

\paragraph{Ablation on key design choices.}
\looseness=-1
We ablate each component of \ours and compare against standard distillation to validate our design choices.
\Cref{tab:ablation-design} isolates the contribution of each component of
\ours. (1) SFT alone improves performance but doesn't fully close the gap, confirming that grounded deferral behavior is helpful but not sufficient on its own.
(2) For \abbr, an off-policy variant, where intermediate refusals are drawn from a fixed corpus rather than from the student's own on-policy samples, improves over SFT but is less effective than the on-policy method.
\Cref{tab:ablation-vanilla} shows that the standard distillation where a stronger teacher (\llama) teaches a weaker student (\llamasmall) is ineffective in this setting.
Together, these results validate the motivation and design of \ours.

\section{Related Work}
\label{related}

\looseness=-1
\paragraph{Improving LLMs in Multi-turn Conversation.}
Several lines of work improve LLM behavior in multi-turn conversation.
A line of work generalizes RLHF to the multi-turn setting by optimizing trajectory-level~\citep{shani2024multiturn,zhou2024archer,gao2025regressing,mt_preference,mt_survey} and turn-level~\citep{wu2025collabllm} rewards.
Another line elicits clarification questions when the user query is ambiguous, through prompting~\citep{mu2023clarifygpt, tree_of_clarifications,zhang2025clarify, deng2023rephrase}, SFT~\citep{andukuri2024stargate}, and RL~\citep{clarify_question_for_retrieval,chen2025learning}.
A recent and concurrent line targets the multi-turn underspecification gap.
Non-parametric methods manipulate the context that reaches the model at inference time~\citep{khalid2025ergo,nanjundappa2025context,liu2026intent,singh2026mt}, but the model's underlying behavior is unchanged, results are sensitive to prompt formulation, and they incur additional latency and orchestration overhead.
RL-based methods apply GRPO to optimize the model~\citep{li2026mitigating, chen2026breaking}, but face the well-known challenges of multi-turn RL: computationally heavy full-conversation rollouts at each step and a sparse end-of-conversation scalar reward that complicates credit assignment~\citep{abdulhai2025lmrl,lu2025onpolicydistillation}.

\looseness=-1
\paragraph{Knowledge distillation.} Knowledge distillation produces meaningful learning signals by exploiting an asymmetry between teacher and student.
The typical form is a \emph{capacity asymmetry}, where a stronger teacher transfers expertise to a weaker student~\citep{hinton2015distilling}. Early work often trains on fixed off-policy datasets~\citep{kim2016sequence,sanh2019distilbert,xie2020self}, while recent work shifts to on-policy distillation, where the student learns from the teacher on its own generations to mitigate the distribution shift~\citep{agarwal2024policy, gu2024minillm, yang2025qwen3,guo2025deepseek}.
The teacher can also be a later checkpoint of the  student~\citep{furlanello2018born, zhang2019your}.
Self-distillation relaxes the capacity requirement by reusing a single model in two roles, but it still depends on an \emph{information asymmetry} that provides the teacher with additional context information, and then trains the student to reproduce the teacher's predictions without it.
This mechanism reliably compresses demonstrated behaviors~\citep{snell2022learning,bai2022constitutional,yang2024self} and factual content~\citep{kujanpaaefficient,eyuboglu2025cartridges} into model weights.
Recent work extends this by conditioning the teacher on external feedback~\citep{scheurer2023training,chen2025retrospective,hubotter2026reinforcement,song2026expanding} or expert demonstrations~\citep{zhao2026self,shenfeld2026self}.
Our \abbr introduces a new axis: \emph{input-view asymmetry}. Teacher and student share the same backbone, parameters, and task information, differing only in how that information is presented. Unlike prior axes, which all rely on external privileged information unavailable in our setting, \abbr requires none: the gap we close is behavioral, not capability-bound.

\section{Conclusion, Limitations, and Future Work}
\label{conclusion}
\looseness=-1
We introduce \name (\ours), a training framework that recovers an LLM's native single-turn competence in underspecified multi-turn conversations. \ours combines two complementary stages: an SFT warm-start that instills grounded deferral and clarification at intermediate turns, and \method~(\abbr) that aligns a student's multi-turn-view distribution with the teacher's information-equivalent single-turn-view distribution, with both instantiated from the same backbone. \abbr requires neither the stronger external teacher that conventional distillation depends on, nor the privileged information that prior self-distillation relies on — both unavailable in this setting. Across five diverse backbones (Llama, Qwen, Phi, OLMo; 3B–14B), \ours attains $92$--$100\%$ of single-turn performance under multi-turn delivery, recovering $75$--$100\%$ of the multi-turn gap, while largely preserving single-turn accuracy and producing more efficient and helpful conversations. The resulting checkpoints further transfer to new domains with minimal adaptation.

\paragraph{Limitations and future work.}
\looseness=-1
\ours isolates one axis of multi-turn difficulty, underspecification, and several realistic dynamics lie outside this scope.
First, the user simulator is cooperative and non-pivoting: it never changes sub-goals or corrects earlier information, so our claims concern the underspecification axis with cooperative users.
We conduct a preliminary stress test where the simulated user states false values and later corrects them: the \ours-trained \llama keeps accuracy similar to that on the same problems with a cooperative user (\Cref{app:noncoop}), so the gain does not appear to hinge on strict cooperation. A systematic evaluation of non-cooperative users is nonetheless beyond our scope.
Second, recovery is measured on the $\sim$100-instance test split with 10 simulations per instance, and we observe consistent trends across different backbones, but a larger suite would strengthen the multi-turn-recovery claim. We view \ours not as a complete solution to multi-turn brittleness, but as evidence that this prevalent failure mode is largely closeable with the model's own single-turn behavior as the only answer supervision.

\begin{ack}
We gratefully acknowledge the support of
NSF under Nos. CCF-1918940 (Expeditions), DMS-2327709 (IHBEM), IIS-2403318 (III);
NIH under No. 1U24NS146314-01,
Gates Foundation,
Stanford Data Applications Initiative,
Wu Tsai Neurosciences Institute,
Stanford Institute for Human-Centered AI,
Allen Institute,
Genentech, SAP, and SCBX.
The content is solely the responsibility of the authors and does not necessarily represent the official views of the funding entities.
This work used Marlowe~\citep{marlowe2025}, Stanford University's GPU-based Computational Instrument, supported by Stanford HAI and Stanford Research Computing.
\end{ack}

\bibliographystyle{plainnat}

\begin{thebibliography}{75}
\providecommand{\natexlab}[1]{#1}
\providecommand{\url}[1]{\texttt{#1}}
\expandafter\ifx\csname urlstyle\endcsname\relax
  \providecommand{\doi}[1]{doi: #1}\else
  \providecommand{\doi}{doi: \begingroup \urlstyle{rm}\Url}\fi

\bibitem[Abdin et~al.(2024)Abdin, Aneja, Behl, Bubeck, Eldan, Gunasekar,
  Harrison, Hewett, Javaheripi, Kauffmann, et~al.]{phi4}
Marah Abdin, Jyoti Aneja, Harkirat Behl, S{\'e}bastien Bubeck, Ronen Eldan,
  Suriya Gunasekar, Michael Harrison, Russell~J Hewett, Mojan Javaheripi, Piero
  Kauffmann, et~al.
\newblock Phi-4 technical report.
\newblock \emph{arXiv preprint arXiv:2412.08905}, 2024.

\bibitem[Abdulhai et~al.(2025)Abdulhai, White, Snell, Sun, Hong, Zhai, Xu, and
  Levine]{abdulhai2025lmrl}
Marwa Abdulhai, Isadora White, Charlie~Victor Snell, Charles Sun, Joey Hong,
  Yuexiang Zhai, Kelvin Xu, and Sergey Levine.
\newblock Lmrl gym: Benchmarks for multi-turn reinforcement learning with
  language models.
\newblock In \emph{International Conference on Machine Learning}, pages
  126--153. PMLR, 2025.

\bibitem[Agarwal et~al.(2024)Agarwal, Vieillard, Zhou, Stanczyk, Garea, Geist,
  and Bachem]{agarwal2024policy}
Rishabh Agarwal, Nino Vieillard, Yongchao Zhou, Piotr Stanczyk, Sabela~Ramos
  Garea, Matthieu Geist, and Olivier Bachem.
\newblock On-policy distillation of language models: Learning from
  self-generated mistakes.
\newblock In \emph{The twelfth international conference on learning
  representations}, 2024.

\bibitem[Andukuri et~al.(2024)Andukuri, Fr{\"a}nken, Gerstenberg, and
  Goodman]{andukuri2024stargate}
Chinmaya Andukuri, Jan-Philipp Fr{\"a}nken, Tobias Gerstenberg, and Noah
  Goodman.
\newblock {ST}ar-{GATE}: Teaching language models to ask clarifying questions.
\newblock In \emph{First Conference on Language Modeling}, 2024.
\newblock URL \url{https://openreview.net/forum?id=CrzAj0kZjR}.

\bibitem[Anthropic(2023)]{Claude}
Anthropic.
\newblock Introducing {Claude}, 2023.
\newblock URL \url{https://www.anthropic.com/index/introducing-claude/}.

\bibitem[Bai et~al.(2022)Bai, Kadavath, Kundu, Askell, Kernion, Jones, Chen,
  Goldie, Mirhoseini, McKinnon, et~al.]{bai2022constitutional}
Yuntao Bai, Saurav Kadavath, Sandipan Kundu, Amanda Askell, Jackson Kernion,
  Andy Jones, Anna Chen, Anna Goldie, Azalia Mirhoseini, Cameron McKinnon,
  et~al.
\newblock Constitutional ai: Harmlessness from ai feedback.
\newblock \emph{arXiv preprint arXiv:2212.08073}, 2022.

\bibitem[Chen et~al.(2025{\natexlab{a}})Chen, Sun, Pfister, and
  Arik]{chen2025learning}
Maximillian Chen, Ruoxi Sun, Tomas Pfister, and Sercan Arik.
\newblock Learning to clarify: Multi-turn conversations with action-based
  contrastive self-training.
\newblock In \emph{International Conference on Learning Representations},
  volume 2025, pages 32244--32279, 2025{\natexlab{a}}.

\bibitem[Chen et~al.(2025{\natexlab{b}})Chen, Kanatsoulis, and
  Leskovec]{chen2025relgnn}
Tianlang Chen, Charilaos Kanatsoulis, and Jure Leskovec.
\newblock {R}el{GNN}: Composite message passing for relational deep learning.
\newblock In \emph{International Conference on Machine Learning}, pages
  8296--8312. PMLR, 2025{\natexlab{b}}.

\bibitem[Chen et~al.(2025{\natexlab{c}})Chen, Xu, Leskovec, and
  Ermon]{chen2025rfg}
Tianlang Chen, Minkai Xu, Jure Leskovec, and Stefano Ermon.
\newblock {RFG}: Test-time scaling for diffusion large language model reasoning
  with reward-free guidance.
\newblock \emph{arXiv preprint arXiv:2509.25604}, 2025{\natexlab{c}}.

\bibitem[Chen et~al.(2025{\natexlab{d}})Chen, Chen, Tajwar, Zhu, Duan,
  Salakhutdinov, and Schneider]{chen2025retrospective}
Wentse Chen, Jiayu Chen, Fahim Tajwar, Hao Zhu, Xintong Duan, Ruslan
  Salakhutdinov, and Jeff Schneider.
\newblock Retrospective in-context learning for temporal credit assignment with
  large language models.
\newblock In \emph{The Thirty-ninth Annual Conference on Neural Information
  Processing Systems}, 2025{\natexlab{d}}.
\newblock URL \url{https://openreview.net/forum?id=QAVpe6a3rp}.

\bibitem[Chen et~al.(2026)Chen, Zhang, Guo, and Zou]{chen2026breaking}
Xingwu Chen, Zhanqiu Zhang, Yiwen Guo, and Difan Zou.
\newblock Breaking contextual inertia: Reinforcement learning with single-turn
  anchors for stable multi-turn interaction.
\newblock \emph{arXiv preprint arXiv:2603.04783}, 2026.

\bibitem[Chu et~al.(2025)Chu, Zhai, Yang, Tong, Xie, Schuurmans, Le, Levine,
  and Ma]{chu2025sft}
Tianzhe Chu, Yuexiang Zhai, Jihan Yang, Shengbang Tong, Saining Xie, Dale
  Schuurmans, Quoc~V Le, Sergey Levine, and Yi~Ma.
\newblock Sft memorizes, rl generalizes: A comparative study of foundation
  model post-training.
\newblock In \emph{International Conference on Machine Learning}, pages
  10818--10838. PMLR, 2025.

\bibitem[Cobbe et~al.(2021)Cobbe, Kosaraju, Bavarian, Chen, Jun, Kaiser,
  Plappert, Tworek, Hilton, Nakano, Hesse, and Schulman]{cobbe2021gsm8k}
Karl Cobbe, Vineet Kosaraju, Mohammad Bavarian, Mark Chen, Heewoo Jun, Lukasz
  Kaiser, Matthias Plappert, Jerry Tworek, Jacob Hilton, Reiichiro Nakano,
  Christopher Hesse, and John Schulman.
\newblock Training verifiers to solve math word problems.
\newblock \emph{arXiv preprint arXiv:2110.14168}, 2021.

\bibitem[D'Amour et~al.(2022)D'Amour, Heller, Moldovan, Adlam, Alipanahi,
  Beutel, Chen, Deaton, Eisenstein, Hoffman, et~al.]{d2022underspecification}
Alexander D'Amour, Katherine Heller, Dan Moldovan, Ben Adlam, Babak Alipanahi,
  Alex Beutel, Christina Chen, Jonathan Deaton, Jacob Eisenstein, Matthew~D
  Hoffman, et~al.
\newblock Underspecification presents challenges for credibility in modern
  machine learning.
\newblock \emph{Journal of Machine Learning Research}, 23\penalty0
  (226):\penalty0 1--61, 2022.

\bibitem[Deng et~al.(2023)Deng, Zhang, Chen, and Gu]{deng2023rephrase}
Yihe Deng, Weitong Zhang, Zixiang Chen, and Quanquan Gu.
\newblock Rephrase and respond: Let large language models ask better questions
  for themselves.
\newblock \emph{arXiv preprint arXiv:2311.04205}, 2023.

\bibitem[Eyuboglu et~al.(2026)Eyuboglu, Ehrlich, Arora, Guha, Zinsley, Liu,
  Rudra, Zou, Mirhoseini, and Re]{eyuboglu2025cartridges}
Sabri Eyuboglu, Ryan~Saul Ehrlich, Simran Arora, Neel Guha, Dylan Zinsley,
  Emily~Ruoyu Liu, Atri Rudra, James Zou, Azalia Mirhoseini, and Christopher
  Re.
\newblock Cartridges: Lightweight and general-purpose long context
  representations via self-study.
\newblock In \emph{The Fourteenth International Conference on Learning
  Representations}, 2026.
\newblock URL \url{https://openreview.net/forum?id=0k5w8O0SNg}.

\bibitem[Ferreira(2008)]{ferreira2008ambiguity}
Victor~S Ferreira.
\newblock Ambiguity, accessibility, and a division of labor for communicative
  success.
\newblock \emph{Psychology of Learning and motivation}, 49:\penalty0 209--246,
  2008.

\bibitem[Frisson(2009)]{Frisson2009SemanticUI}
Steven Frisson.
\newblock Semantic underspecification in language processing.
\newblock \emph{Lang. Linguistics Compass}, 3:\penalty0 111--127, 2009.
\newblock URL \url{https://api.semanticscholar.org/CorpusID:13384476}.

\bibitem[Furlanello et~al.(2018)Furlanello, Lipton, Tschannen, Itti, and
  Anandkumar]{furlanello2018born}
Tommaso Furlanello, Zachary Lipton, Michael Tschannen, Laurent Itti, and Anima
  Anandkumar.
\newblock Born again neural networks.
\newblock In \emph{International Conference on Machine Learning}, pages
  1607--1616. PMLR, 2018.

\bibitem[Gao et~al.(2025)Gao, Zhan, Chang, Swamy, Brantley, Lee, and
  Sun]{gao2025regressing}
Zhaolin Gao, Wenhao Zhan, Jonathan~Daniel Chang, Gokul Swamy, Kiant{\'e}
  Brantley, Jason~D. Lee, and Wen Sun.
\newblock Regressing the relative future: Efficient policy optimization for
  multi-turn {RLHF}.
\newblock In \emph{The Thirteenth International Conference on Learning
  Representations}, 2025.
\newblock URL \url{https://openreview.net/forum?id=cVyELMpMRS}.

\bibitem[Geirhos et~al.(2020)Geirhos, Jacobsen, Michaelis, Zemel, Brendel,
  Bethge, and Wichmann]{geirhos2020shortcut}
Robert Geirhos, J{\"o}rn-Henrik Jacobsen, Claudio Michaelis, Richard Zemel,
  Wieland Brendel, Matthias Bethge, and Felix~A Wichmann.
\newblock Shortcut learning in deep neural networks.
\newblock \emph{Nature Machine Intelligence}, 2\penalty0 (11):\penalty0
  665--673, 2020.

\bibitem[Grattafiori et~al.(2024)Grattafiori, Dubey, Jauhri, Pandey, Kadian,
  Al-Dahle, Letman, Mathur, Schelten, Vaughan, et~al.]{llama3}
Aaron Grattafiori, Abhimanyu Dubey, Abhinav Jauhri, Abhinav Pandey, Abhishek
  Kadian, Ahmad Al-Dahle, Aiesha Letman, Akhil Mathur, Alan Schelten, Alex
  Vaughan, et~al.
\newblock The llama 3 herd of models.
\newblock \emph{arXiv preprint arXiv:2407.21783}, 2024.

\bibitem[Gu et~al.(2024)Gu, Dong, Wei, and Huang]{gu2024minillm}
Yuxian Gu, Li~Dong, Furu Wei, and Minlie Huang.
\newblock Minillm: Knowledge distillation of large language models.
\newblock In \emph{The twelfth international conference on learning
  representations}, 2024.

\bibitem[Guo et~al.(2025)Guo, Yang, Zhang, Song, Zhang, Xu, Zhu, Ma, Wang, Bi,
  et~al.]{guo2025deepseek}
Daya Guo, Dejian Yang, Haowei Zhang, Junxiao Song, Ruoyu Zhang, Runxin Xu,
  Qihao Zhu, Shirong Ma, Peiyi Wang, Xiao Bi, et~al.
\newblock Deepseek-r1: Incentivizing reasoning capability in llms via
  reinforcement learning.
\newblock \emph{arXiv preprint arXiv:2501.12948}, 2025.

\bibitem[Herlihy et~al.(2024)Herlihy, Neville, Schnabel, and
  Swaminathan]{herlihy2024overcoming}
Christine Herlihy, Jennifer Neville, Tobias Schnabel, and Adith Swaminathan.
\newblock On overcoming miscalibrated conversational priors in {LLM}-based
  chatbots.
\newblock In \emph{The 40th Conference on Uncertainty in Artificial
  Intelligence}, 2024.
\newblock URL \url{https://openreview.net/forum?id=iXQIgIsmIr}.

\bibitem[Hinton et~al.(2015)Hinton, Vinyals, and Dean]{hinton2015distilling}
Geoffrey Hinton, Oriol Vinyals, and Jeff Dean.
\newblock Distilling the knowledge in a neural network.
\newblock \emph{arXiv preprint arXiv:1503.02531}, 2015.

\bibitem[Hu et~al.(2022)Hu, Shen, Wallis, Allen{-}Zhu, Li, Wang, Wang, and
  Chen]{lora}
Edward~J. Hu, Yelong Shen, Phillip Wallis, Zeyuan Allen{-}Zhu, Yuanzhi Li,
  Shean Wang, Lu~Wang, and Weizhu Chen.
\newblock Lora: Low-rank adaptation of large language models.
\newblock In \emph{ICLR}, 2022.

\bibitem[Huang et~al.(2024)Huang, Chen, Mishra, Zheng, Yu, Song, and
  Zhou]{huang2023large}
Jie Huang, Xinyun Chen, Swaroop Mishra, Huaixiu~Steven Zheng, Adams~Wei Yu,
  Xinying Song, and Denny Zhou.
\newblock Large language models cannot self-correct reasoning yet.
\newblock In \emph{The Twelfth International Conference on Learning
  Representations}, 2024.
\newblock URL \url{https://openreview.net/forum?id=IkmD3fKBPQ}.

\bibitem[H\"{u}botter et~al.(2026)H\"{u}botter, L\"{u}beck, Behric, Baumann,
  Bagatella, Marta, Hakimi, Shenfeld, Kleine~Buening, Guestrin, and
  Krause]{hubotter2026reinforcement}
Jonas H\"{u}botter, Frederike L\"{u}beck, Lejs~Deen Behric, Anton Baumann,
  Marco Bagatella, Daniel Marta, Ido Hakimi, Idan Shenfeld, Thomas
  Kleine~Buening, Carlos Guestrin, and Andreas Krause.
\newblock Reinforcement learning via self-distillation.
\newblock In \emph{International Conference on Machine Learning}, pages
  48677--48721. PMLR, 2026.

\bibitem[Kapfer et~al.(2025)Kapfer, Stine, Narasimhan, Mentzel, and
  Cand{\`e}s]{marlowe2025}
Craig Kapfer, Kurt Stine, Balasubramanian Narasimhan, Christopher Mentzel, and
  Emmanuel Cand{\`e}s.
\newblock {Marlowe: Stanford's GPU-based Computational Instrument}, 2025.
\newblock URL \url{https://doi.org/10.5281/zenodo.14751899}.

\bibitem[Khalid et~al.(2025)Khalid, Jeyaganthan, Do, Fu, Sharma, O’Brien, and
  Zhu]{khalid2025ergo}
Haziq~Mohammad Khalid, Athikash Jeyaganthan, Timothy Do, Yicheng Fu, Vasu
  Sharma, Sean O’Brien, and Kevin Zhu.
\newblock Ergo: Entropy-guided resetting for generation optimization in
  multi-turn language models.
\newblock In \emph{Proceedings of the 2nd Workshop on Uncertainty-Aware NLP
  (UncertaiNLP 2025)}, pages 273--286, 2025.

\bibitem[Kim et~al.(2023)Kim, Kim, Jeon, Park, and
  Kang]{tree_of_clarifications}
Gangwoo Kim, Sungdong Kim, Byeongguk Jeon, Joonsuk Park, and Jaewoo Kang.
\newblock Tree of clarifications: Answering ambiguous questions with
  retrieval-augmented large language models.
\newblock In \emph{EMNLP}, 2023.

\bibitem[Kim and Rush(2016)]{kim2016sequence}
Yoon Kim and Alexander~M Rush.
\newblock Sequence-level knowledge distillation.
\newblock In \emph{Proceedings of the 2016 conference on empirical methods in
  natural language processing}, pages 1317--1327, 2016.

\bibitem[Kujanp{\"a}{\"a} et~al.(2025)Kujanp{\"a}{\"a}, Marttinen, Valpola, and
  Ilin]{kujanpaaefficient}
Kalle Kujanp{\"a}{\"a}, Pekka Marttinen, Harri Valpola, and Alexander Ilin.
\newblock Efficient knowledge injection in {LLM}s via self-distillation.
\newblock \emph{Transactions on Machine Learning Research}, 2025.
\newblock ISSN 2835-8856.
\newblock URL \url{https://openreview.net/forum?id=drYpdSnRJk}.

\bibitem[Laban et~al.(2026)Laban, Hayashi, Zhou, and Neville]{laban2025llms}
Philippe Laban, Hiroaki Hayashi, Yingbo Zhou, and Jennifer Neville.
\newblock {LLM}s get lost in multi-turn conversation.
\newblock In \emph{The Fourteenth International Conference on Learning
  Representations}, 2026.
\newblock URL \url{https://openreview.net/forum?id=VKGTGGcwl6}.

\bibitem[Lampinen et~al.(2025)Lampinen, Chaudhry, Chan, Wild, Wan, Ku,
  Bornschein, Pascanu, Shanahan, and McClelland]{lampinen2025generalization}
Andrew~K Lampinen, Arslan Chaudhry, Stephanie~CY Chan, Cody Wild, Diane Wan,
  Alex Ku, J{\"o}rg Bornschein, Razvan Pascanu, Murray Shanahan, and James~L
  McClelland.
\newblock On the generalization of language models from in-context learning and
  finetuning: a controlled study.
\newblock \emph{arXiv preprint arXiv:2505.00661}, 2025.

\bibitem[Li et~al.(2026)Li, Chen, Zhang, Yang, Zhang, Li, Cao, Zeng, Wu, Jiang,
  et~al.]{li2026mitigating}
Ming Li, Pei Chen, Zhenhao Zhang, Tao Yang, Xinyang Zhang, Han Li, Tianyu Cao,
  Ming Zeng, Zhuofeng Wu, Meng Jiang, et~al.
\newblock Mitigating lost in multi-turn conversation via curriculum rl with
  verifiable accuracy and abstention rewards.
\newblock In \emph{Proceedings of the 64th Annual Meeting of the Association
  for Computational Linguistics (Volume 1: Long Papers)}, pages 33364--33381,
  2026.

\bibitem[Lin(1991)]{jsdloss}
Jianhua Lin.
\newblock Divergence measures based on the shannon entropy.
\newblock \emph{IEEE Transactions on Information theory}, 37\penalty0
  (1):\penalty0 145--151, 1991.

\bibitem[Liu et~al.(2026)Liu, Zhu, Feng, Ma, Wang, and Meng]{liu2026intent}
Geng Liu, Fei Zhu, Rong Feng, Changyi Ma, Shiqi Wang, and Gaofeng Meng.
\newblock Intent mismatch causes llms to get lost in multi-turn conversation.
\newblock \emph{arXiv preprint arXiv:2602.07338}, 2026.

\bibitem[Liu et~al.(2025)Liu, Chen, Lu, Ye, Chen, Xing, and
  Zou]{liu2025fractional}
Sheng Liu, Tianlang Chen, Pan Lu, Haotian Ye, Yizheng Chen, Lei Xing, and James
  Zou.
\newblock Fractional reasoning via latent steering vectors improves inference
  time compute.
\newblock \emph{arXiv preprint arXiv:2506.15882}, 2025.

\bibitem[Lu and Lab(2025)]{lu2025onpolicydistillation}
Kevin Lu and Thinking~Machines Lab.
\newblock On-policy distillation.
\newblock \emph{Thinking Machines Lab: Connectionism}, 2025.
\newblock \doi{10.64434/tml.20251026}.
\newblock https://thinkingmachines.ai/blog/on-policy-distillation.

\bibitem[Lu et~al.(2025)Lu, Sun, Du, Ling, Yao, Liu, and Chen]{lu2025scaling}
Miao Lu, Weiwei Sun, Weihua Du, Zhan Ling, Xuesong Yao, Kang Liu, and Jiecao
  Chen.
\newblock Scaling llm multi-turn rl with end-to-end summarization-based context
  management.
\newblock \emph{arXiv preprint arXiv:2510.06727}, 2025.

\bibitem[Mu et~al.(2023{\natexlab{a}})Mu, Shi, Wang, Yu, Zhang, Wang, Liu, and
  Wang]{mu2023clarifygpt}
Fangwen Mu, Lin Shi, Song Wang, Zhuohao Yu, Binquan Zhang, Chenxue Wang,
  Shichao Liu, and Qing Wang.
\newblock Clarifygpt: Empowering llm-based code generation with intention
  clarification.
\newblock \emph{arXiv preprint arXiv:2310.10996}, 2023{\natexlab{a}}.

\bibitem[Mu et~al.(2023{\natexlab{b}})Mu, Li, and Goodman]{mu2023learning}
Jesse Mu, Xiang Li, and Noah Goodman.
\newblock Learning to compress prompts with gist tokens.
\newblock \emph{Advances in Neural Information Processing Systems},
  36:\penalty0 19327--19352, 2023{\natexlab{b}}.

\bibitem[Nanjundappa and Maaheshwari(2025)]{nanjundappa2025context}
Bhargav~Chickmagalur Nanjundappa and Spandan Maaheshwari.
\newblock Context branching for llm conversations: A version control approach
  to exploratory programming.
\newblock \emph{arXiv preprint arXiv:2512.13914}, 2025.

\bibitem[OLMo et~al.(2024)OLMo, Walsh, Soldaini, Groeneveld, Lo, Arora, Bhagia,
  Gu, Huang, Jordan, et~al.]{olmo2}
Team OLMo, Pete Walsh, Luca Soldaini, Dirk Groeneveld, Kyle Lo, Shane Arora,
  Akshita Bhagia, Yuling Gu, Shengyi Huang, Matt Jordan, et~al.
\newblock 2 olmo 2 furious.
\newblock \emph{arXiv preprint arXiv:2501.00656}, 2024.

\bibitem[OpenAI(2023)]{GPT-4}
OpenAI.
\newblock Gpt-4 technical report.
\newblock \emph{arXiv preprint arXiv:2303.08774}, 2023.

\bibitem[Sanh et~al.(2019)Sanh, Debut, Chaumond, and Wolf]{sanh2019distilbert}
Victor Sanh, Lysandre Debut, Julien Chaumond, and Thomas Wolf.
\newblock Distilbert, a distilled version of bert: smaller, faster, cheaper and
  lighter.
\newblock \emph{arXiv preprint arXiv:1910.01108}, 2019.

\bibitem[Scheurer et~al.(2023)Scheurer, Campos, Korbak, Chan, Chen, Cho, and
  Perez]{scheurer2023training}
J{\'e}r{\'e}my Scheurer, Jon~Ander Campos, Tomasz Korbak, Jun~Shern Chan,
  Angelica Chen, Kyunghyun Cho, and Ethan Perez.
\newblock Training language models with language feedback at scale.
\newblock \emph{arXiv preprint arXiv:2303.16755}, 2023.

\bibitem[Schick et~al.(2023)Schick, Dwivedi-Yu, Dess{\`\i}, Raileanu, Lomeli,
  Hambro, Zettlemoyer, Cancedda, and Scialom]{schick2023toolformer}
Timo Schick, Jane Dwivedi-Yu, Roberto Dess{\`\i}, Roberta Raileanu, Maria
  Lomeli, Eric Hambro, Luke Zettlemoyer, Nicola Cancedda, and Thomas Scialom.
\newblock Toolformer: Language models can teach themselves to use tools.
\newblock \emph{Advances in neural information processing systems},
  36:\penalty0 68539--68551, 2023.

\bibitem[Sclar et~al.(2024)Sclar, Choi, Tsvetkov, and
  Suhr]{sclar2024quantifying}
Melanie Sclar, Yejin Choi, Yulia Tsvetkov, and Alane Suhr.
\newblock Quantifying language models' sensitivity to spurious features in
  prompt design or: How i learned to start worrying about prompt formatting.
\newblock In \emph{International Conference on Learning Representations},
  volume 2024, pages 25055--25083, 2024.

\bibitem[Shani et~al.(2024)Shani, Rosenberg, Cassel, Lang, Calandriello,
  Zipori, Noga, Keller, Piot, Szpektor, Hassidim, Matias, and
  Munos]{shani2024multiturn}
Lior Shani, Aviv Rosenberg, Asaf Cassel, Oran Lang, Daniele Calandriello,
  Avital Zipori, Hila Noga, Orgad Keller, Bilal Piot, Idan Szpektor, Avinatan
  Hassidim, Yossi Matias, and Remi Munos.
\newblock Multi-turn reinforcement learning with preference human feedback.
\newblock In \emph{The Thirty-eighth Annual Conference on Neural Information
  Processing Systems}, 2024.
\newblock URL \url{https://openreview.net/forum?id=rVSc3HIZS4}.

\bibitem[Sharma et~al.(2024)Sharma, Tong, Korbak, Duvenaud, Askell, Bowman,
  Durmus, Hatfield-Dodds, Johnston, Kravec, et~al.]{sharma2024towards}
Mrinank Sharma, Meg Tong, Tomek Korbak, David Duvenaud, Amanda Askell, Sam
  Bowman, Esin Durmus, Zac Hatfield-Dodds, Scott Johnston, Shauna Kravec,
  et~al.
\newblock Towards understanding sycophancy in language models.
\newblock In \emph{International Conference on Learning Representations},
  volume 2024, pages 110--144, 2024.

\bibitem[Shenfeld et~al.(2026)Shenfeld, Damani, H\"{u}botter, and
  Agrawal]{shenfeld2026self}
Idan Shenfeld, Mehul Damani, Jonas H\"{u}botter, and Pulkit Agrawal.
\newblock Self-distillation enables continual learning.
\newblock In \emph{International Conference on Machine Learning}, pages
  111542--111558. PMLR, 2026.

\bibitem[Shi et~al.(2024)Shi, Yuan, Wu, Wang, and Feng]{mt_preference}
Wentao Shi, Mengqi Yuan, Junkang Wu, Qifan Wang, and Fuli Feng.
\newblock Direct multi-turn preference optimization for language agents.
\newblock In \emph{EMNLP}. Association for Computational Linguistics, 2024.

\bibitem[Singh et~al.(2026)Singh, Tu, Ballesteros, Sun, Ghoshal, Yuan,
  Benajiba, Ravi, and Roth]{singh2026mt}
Jyotika Singh, Fang Tu, Miguel Ballesteros, Weiyi Sun, Sandip Ghoshal, Michelle
  Yuan, Yassine Benajiba, Sujith Ravi, and Dan Roth.
\newblock Mt-osc: Path for llms that get lost in multi-turn conversation.
\newblock \emph{arXiv preprint arXiv:2604.08782}, 2026.

\bibitem[Snell et~al.(2022)Snell, Klein, and Zhong]{snell2022learning}
Charlie Snell, Dan Klein, and Ruiqi Zhong.
\newblock Learning by distilling context.
\newblock \emph{arXiv preprint arXiv:2209.15189}, 2022.

\bibitem[Snell et~al.(2025)Snell, Lee, Xu, and Kumar]{snell2025scaling}
Charlie~Victor Snell, Jaehoon Lee, Kelvin Xu, and Aviral Kumar.
\newblock Scaling {LLM} test-time compute optimally can be more effective than
  scaling parameters for reasoning.
\newblock In \emph{The Thirteenth International Conference on Learning
  Representations}, 2025.
\newblock URL \url{https://openreview.net/forum?id=4FWAwZtd2n}.

\bibitem[Song et~al.(2026)Song, Chen, Tajwar, Munos, Pathak, Bagnell, Singh,
  and Zanette]{song2026expanding}
Yuda Song, Lili Chen, Fahim Tajwar, R\'{e}mi Munos, Deepak Pathak, Drew
  Bagnell, Aarti Singh, and Andrea Zanette.
\newblock Expanding the capabilities of reinforcement learning via text
  feedback.
\newblock In \emph{International Conference on Machine Learning}, pages
  114482--114522. PMLR, 2026.

\bibitem[Team(2024)]{qwen2.5}
Qwen Team.
\newblock Qwen2.5: A party of foundation models, September 2024.
\newblock URL \url{https://qwenlm.github.io/blog/qwen2.5/}.

\bibitem[Verma et~al.(2024)Verma, Bhambri, and Kambhampati]{verma2024brittle}
Mudit Verma, Siddhant Bhambri, and Subbarao Kambhampati.
\newblock On the brittle foundations of react prompting for agentic large
  language models.
\newblock \emph{arXiv preprint arXiv:2405.13966}, 2024.

\bibitem[Wu et~al.(2025{\natexlab{a}})Wu, Galley, Peng, Cheng, Li, Dou, Cai,
  Zou, Leskovec, and Gao]{wu2025collabllm}
Shirley Wu, Michel Galley, Baolin Peng, Hao Cheng, Gavin Li, Yao Dou, Weixin
  Cai, James Zou, Jure Leskovec, and Jianfeng Gao.
\newblock Collabllm: From passive responders to active collaborators.
\newblock In \emph{International Conference on Machine Learning}, pages
  67260--67283. PMLR, 2025{\natexlab{a}}.

\bibitem[Wu et~al.(2025{\natexlab{b}})Wu, Sun, Li, Welleck, and
  Yang]{wu2025inference}
Yangzhen Wu, Zhiqing Sun, Shanda Li, Sean Welleck, and Yiming Yang.
\newblock Inference scaling laws: An empirical analysis of compute-optimal
  inference for {LLM} problem-solving.
\newblock In \emph{The Thirteenth International Conference on Learning
  Representations}, 2025{\natexlab{b}}.
\newblock URL \url{https://openreview.net/forum?id=VNckp7JEHn}.

\bibitem[Xie et~al.(2020)Xie, Luong, Hovy, and Le]{xie2020self}
Qizhe Xie, Minh-Thang Luong, Eduard Hovy, and Quoc~V Le.
\newblock Self-training with noisy student improves imagenet classification.
\newblock In \emph{Proceedings of the IEEE/CVF conference on computer vision
  and pattern recognition}, pages 10687--10698, 2020.

\bibitem[Yan et~al.(2024)Yan, Mao, Ji, Zhang, Patil, Stoica, and
  Gonzalez]{2024bfcl}
Fanjia Yan, Huanzhi Mao, Charlie Cheng-Jie Ji, Tianjun Zhang, Shishir~G. Patil,
  Ion Stoica, and Joseph~E. Gonzalez.
\newblock Berkeley function calling leaderboard.
\newblock
  \url{https://gorilla.cs.berkeley.edu/blogs/8_berkeley_function_calling_leaderboard.html},
  2024.

\bibitem[Yang et~al.(2025)Yang, Li, Yang, Zhang, Hui, Zheng, Yu, Gao, Huang,
  Lv, et~al.]{yang2025qwen3}
An~Yang, Anfeng Li, Baosong Yang, Beichen Zhang, Binyuan Hui, Bo~Zheng, Bowen
  Yu, Chang Gao, Chengen Huang, Chenxu Lv, et~al.
\newblock Qwen3 technical report.
\newblock \emph{arXiv preprint arXiv:2505.09388}, 2025.

\bibitem[Yang et~al.(2024)Yang, Pang, Feng, Wang, Chen, Zhu, and
  Liu]{yang2024self}
Zhaorui Yang, Tianyu Pang, Haozhe Feng, Han Wang, Wei Chen, Minfeng Zhu, and
  Qian Liu.
\newblock Self-distillation bridges distribution gap in language model
  fine-tuning.
\newblock In \emph{Proceedings of the 62nd Annual Meeting of the Association
  for Computational Linguistics (Volume 1: Long Papers)}, pages 1028--1043,
  2024.

\bibitem[Yu et~al.(2018)Yu, Zhang, Yang, Yasunaga, Wang, Li, Ma, Li, Yao,
  Roman, et~al.]{yu2018spider}
Tao Yu, Rui Zhang, Kai Yang, Michihiro Yasunaga, Dongxu Wang, Zifan Li, James
  Ma, Irene Li, Qingning Yao, Shanelle Roman, et~al.
\newblock Spider: A large-scale human-labeled dataset for complex and
  cross-domain semantic parsing and text-to-sql task.
\newblock In \emph{Proceedings of the 2018 conference on empirical methods in
  natural language processing}, pages 3911--3921, 2018.

\bibitem[Zamani et~al.(2020)Zamani, Dumais, Craswell, Bennett, and
  Lueck]{clarify_question_for_retrieval}
Hamed Zamani, Susan~T. Dumais, Nick Craswell, Paul~N. Bennett, and Gord Lueck.
\newblock Generating clarifying questions for information retrieval.
\newblock In \emph{WWW}, 2020.

\bibitem[Zhang et~al.(2025)Zhang, Dai, Wu, Yang, Wang, Tang, and
  Liu]{mt_survey}
Chen Zhang, Xinyi Dai, Yaxiong Wu, Qu~Yang, Yasheng Wang, Ruiming Tang, and
  Yong Liu.
\newblock A survey on multi-turn interaction capabilities of large language
  models.
\newblock \emph{arXiv preprint arXiv:2501.09959}, 2025.

\bibitem[Zhang et~al.(2019)Zhang, Song, Gao, Chen, Bao, and Ma]{zhang2019your}
Linfeng Zhang, Jiebo Song, Anni Gao, Jingwei Chen, Chenglong Bao, and Kaisheng
  Ma.
\newblock Be your own teacher: Improve the performance of convolutional neural
  networks via self distillation.
\newblock In \emph{Proceedings of the IEEE/CVF international conference on
  computer vision}, pages 3713--3722, 2019.

\bibitem[Zhang and Choi(2025)]{zhang2025clarify}
Michael~JQ Zhang and Eunsol Choi.
\newblock Clarify when necessary: Resolving ambiguity through interaction with
  lms.
\newblock In \emph{Findings of the Association for Computational Linguistics:
  NAACL 2025}, pages 5526--5543, 2025.

\bibitem[Zhao et~al.(2026)Zhao, Xie, Liu, Huang, Pang, Chen, and
  Grover]{zhao2026self}
Siyan Zhao, Zhihui Xie, Mengchen Liu, Jing Huang, Guan Pang, Feiyu Chen, and
  Aditya Grover.
\newblock Self-distilled reasoner: On-policy self-distillation for large
  language models.
\newblock In \emph{International Conference on Machine Learning}, pages
  162433--162448. PMLR, 2026.

\bibitem[Zhou et~al.(2024)Zhou, Zanette, Pan, Levine, and
  Kumar]{zhou2024archer}
Yifei Zhou, Andrea Zanette, Jiayi Pan, Sergey Levine, and Aviral Kumar.
\newblock {A}r{CH}er: Training language model agents via hierarchical
  multi-turn {RL}.
\newblock In \emph{International Conference on Machine Learning}, pages
  62178--62209. PMLR, 2024.

\bibitem[Zipf(1949)]{zipf1949human}
George~Kingsley Zipf.
\newblock \emph{Human behavior and the principle of least effort: An
  introduction to human ecology}.
\newblock Addison-Wesley Press, 1949.

\end{thebibliography}

\newpage
\appendix

\begin{table}[t]
\centering
\footnotesize
\setlength{\tabcolsep}{4pt}
\caption{Inference-time interventions fail to recover the multi-turn gap.
Each row applies a different lightweight intervention on top of the same baseline model in the \sharded setting; accuracy (\%) is reported on the same evaluation setup as Table~\ref{tab:main}.}
\label{tab:prompt-baselines}
\begin{tabular}{lccccc}
\toprule
Method & \llamasmall & \llama & \qwen & \olmo & \phimodel \\
\midrule
\sharded (no intervention)            & $39.2$ & $44.7$ & $58.9$ & $54.4$ & $61.2$ \\
\midrule
\;\;+ \promptrule                     & $29.4$ & $31.3$ & $54.1$ & $61.3$ & $67.9$ \\
\;\;+ \promptselfcheck                & $27.2$ & $39.9$ & $51.6$ & $56.1$ & $68.2$ \\
\;\;+ \gatedexternal\ (4o-mini)       & $28.5$ & $36.8$ & $45.6$ & $37.9$ & $42.5$ \\
\;\;+ \gatedself                      & $30.7$ & $\phantom{0}4.4$ & $37.7$ & $30.0$ & $63.0$ \\
\midrule
\;\;+ \ours                           & $\bm{60.4}$ & $\bm{71.7}$ & $\bm{78.7}$ & $\bm{77.1}$ & $\bm{82.4}$ \\
\bottomrule
\end{tabular}
\end{table}

\section{Inference-time Methods}
\label{app:test-time}

\looseness=-1
We include four lightweight inference-time intervention baselines that do not modify model weights. Inference-time methods are a cheap alternative to post-training, and recent work shows that model behavior can be substantially improved without changing any parameters~\citep{snell2025scaling,liu2025fractional,chen2025rfg}. The goal is to test whether the multi-turn gap can be closed with training-free methods. The interventions include (i) instructing the model to defer until enough information is present, and (ii) gating the model's answer attempts via a separate judge pass. See~\Cref{app:test-time-prompts} for the exact prompts used.

\paragraph{Prompt-based interventions.}
\promptrule prepends a short behavioral rule to the system prompt instructing the model to defer until every numerical or factual value required to solve the task has been explicitly stated by the user.
\promptselfcheck requires the model to prefix every response with \texttt{READY: YES} or \texttt{READY: NO}, depending on whether it judges the available information sufficient. Only \texttt{YES} responses are scored; \texttt{NO} responses are treated as deferrals.

\paragraph{Judge-gated (two-pass) interventions.}
At every turn, an auxiliary judge first classifies whether the conversation so far contains enough information to attempt an answer. The assistant only generates an answer attempt when the judge returns \texttt{YES}. \gatedexternal uses GPT-4o-mini as the judge, while \gatedself uses a second pass of the assistant model itself with a judge-style system prompt.

\paragraph{Results.}
\Cref{tab:prompt-baselines} reports \sharded accuracy under each intervention. None of the four interventions consistently improves over the no-intervention \sharded baseline; in many model–intervention combinations they actively degrade performance. These findings are consistent with prior observations that prompt-level interventions struggle to overturn ingrained model behavior~\citep{huang2023large,sclar2024quantifying,verma2024brittle}, and they motivate the parametric approach taken by \ours.

\section{Experimental Details}
\label{app:exp-details}

\subsection{Compute Resource}
\label{app:hw}

\looseness=-1
All training and evaluation runs use $8\times$ A100 and H100 GPUs in a single node.

\subsection{Evaluation Protocol}
\label{app:eval}

\looseness=-1
We follow the simulation protocol of~\citet{laban2025llms}. The user simulator is GPT-4o-mini at $T=1.0$, with access to the full sharded instruction. At the first turn the user simulator always reveals the general goal (e.g., ``How much \dots?'') and in subsequent turns it selects the next shard to reveal conditioned on the conversation history, mirroring how a real user would naturally introduce missing details. Assistant temperature is also fixed at $T=1.0$. A strategy classifier categorizes each assistant response into one of seven types (clarification, refusal, hedging, interrogation, discussion, missing, or answer attempt); only answer attempts are scored. If an answer attempt is incorrect, the simulator continues revealing shards until a correct attempt is produced or the shard set is exhausted. We run $n=10$ simulations per instance and report mean accuracy.

\subsection{Multi-turn Corpus Construction}
\label{app:corpus}

\paragraph{Source instructions and candidate filtering.}
\looseness=-1
Training shardings are produced from GSM8K~\citep{cobbe2021gsm8k} instructions disjoint from the $103$-instance test split released by~\citet{laban2025llms}. We pre-filter to questions with at least three sentences, ensuring each problem can yield a non-trivial multi-shard decomposition.

\paragraph{Sharding pipeline.}
\looseness=-1
Each candidate is processed by the two-step ``lazification'' pipeline of~\citet{laban2025llms}: GPT-4o first segments the question into atomic facts, then rephrases each segment into a conversational shard $\{u_i\}$. The goal-revealing shard $u_1$ (``how many \dots ?'') is always placed in the first turn; the remaining shards carry the supporting facts.

\paragraph{Verification filter.}
\looseness=-1
A candidate is admitted only if every base model in our experiment satisfies the information-equivalence criterion empirically: $\bar P(\pi; q) > 0$, $\bar P(\pi; q_{\text{concat}}) \ge \tau\, \bar P(\pi; q)$, and the same relation under a shuffled-concat presentation where the shards are randomly shuffled before concatenation.  We restrict to shard counts $N \in [3, 8]$.

\paragraph{Intermediate-turn supervision (refusals).}
\looseness=-1
Intermediate assistant turns $a_t$ ($t<N$) are synthesized by GPT-4o-mini, where roughly $45\%$ of deferrals are phrased as questions and $55\%$ as statements. Specific clarifying questions are retained only in the later turns, and \emph{only} when a GPT-4o-mini judge verifies that the request actually targets information that will be revealed in the unrevealed shards.

\paragraph{Final-turn supervision.}
\looseness=-1
For each model and each problem we run $n=10$ \concat rollouts at $T=1.0$ and top-$p = 0.9$, and retain only the correct generations. One correct rollout is paired with each shard-order variant per epoch, so the final-turn target rotates across epochs (mirroring the stylistic variation applied to intermediate turns). When a problem has no correct \concat rollout for a particular backbone (a small minority of records), we fall back to the GSM8K reference solution. This makes the SFT signal \emph{model-specific}: each backbone is supervised on its own solution traces, keeping the SFT distribution close to the model's prior.

\subsection{Hyperparameters}
\label{app:sft-hp}

\looseness=-1
LoRA targets are $\{q,k,v,o\}_{\text{proj}} \cup \{\text{gate},\text{up},\text{down}\}_{\text{proj}}$ for \llamasmall, \llama, \qwen, and \olmo, and the fused $\{qkv_{\text{proj}}, o_{\text{proj}}, \text{gate\_up}_{\text{proj}}, \text{down}_{\text{proj}}\}$ for \phimodel. We set LoRA rank $r$ based on model size: $r=8$ for \llamasmall, $r=16$ for \llama and \qwen, and $r=32$ for \olmo and \phimodel. We hold $\alpha/r = 2$ and LoRA dropout $0.05$ across all rank settings. SFT loss is the standard causal-LM cross-entropy computed only over assistant tokens. The learning rate for SFT is set to 2e-4 for all models. The effective batch size for SFT is $32$ for smaller models (\llamasmall, \llama, and \qwen) and $16$ for larger ones (\olmo and \phimodel). The \abbr LoRA adapter is initialized from the SFT checkpoint of the same backbone. We train with token-level Jensen--Shannon divergence loss for 1 epoch for each model. The learning rate for \abbr is set to 1e-4 for all models. The effective batch size for \abbr is the same as in the SFT stage.

\looseness=-1
Checkpoints are saved at the end of each epoch. SFT checkpoints are selected on a held-out development split, disjoint from both the training and the test split. \abbr is a single-epoch pass with no checkpoint selection. We note that SFT checkpoint candidates perform similarly on the development split, so we additionally evaluate a fixed final SFT checkpoint for every backbone (\Cref{tab:fixedckpt}). Every fixed-checkpoint value lies within the 95\% confidence interval of the dev-selected one, the largest shifts are $+2.6$ and $-1.5$ points, and all conclusions are unchanged. We therefore recommend using the final SFT checkpoint directly.

\begin{table}[t]
\centering
\footnotesize
\setlength{\tabcolsep}{3pt}
\caption{\ours results with the dev-selected SFT checkpoint (as in \Cref{tab:main}) and with a fixed final SFT checkpoint, with 95\% bootstrap confidence intervals. For \llama and \phimodel the dev-selected checkpoint is the final one, so the rows coincide.}
\label{tab:fixedckpt}
\resizebox{\linewidth}{!}{%
\begin{tabular}{llccccc}
\toprule
Setting & Checkpoint & \llamasmall & \llama & \qwen & \olmo & \phimodel \\
\midrule
\multirow{2}{*}{\full}
 & dev-selected & $58.3$ [$51.1$, $65.4$] & $73.8$ [$67.4$, $79.8$] & $83.6$ [$78.2$, $88.5$] & $79.5$ [$73.6$, $85.2$] & $89.8$ [$85.0$, $94.2$] \\
 & fixed        & $60.9$ [$53.9$, $67.6$] & $73.8$ [$67.4$, $79.8$] & $83.9$ [$78.7$, $88.5$] & $80.5$ [$74.8$, $85.6$] & $89.8$ [$85.0$, $94.2$] \\
\midrule
\multirow{2}{*}{\concat}
 & dev-selected & $52.6$ [$45.5$, $59.6$] & $72.3$ [$65.8$, $78.5$] & $82.6$ [$77.3$, $87.5$] & $77.8$ [$71.6$, $83.7$] & $88.0$ [$82.7$, $92.6$] \\
 & fixed        & $54.0$ [$47.2$, $60.7$] & $72.3$ [$65.8$, $78.5$] & $85.2$ [$80.4$, $89.6$] & $78.7$ [$72.8$, $84.2$] & $88.0$ [$82.7$, $92.6$] \\
\midrule
\multirow{2}{*}{\sharded}
 & dev-selected & $60.4$ [$53.6$, $67.2$] & $71.7$ [$65.4$, $77.8$] & $78.7$ [$73.1$, $84.0$] & $77.1$ [$71.4$, $82.5$] & $82.4$ [$77.2$, $87.3$] \\
 & fixed        & $59.9$ [$53.4$, $66.3$] & $71.7$ [$65.4$, $77.8$] & $79.4$ [$74.0$, $84.5$] & $75.6$ [$70.2$, $80.8$] & $82.4$ [$77.2$, $87.3$] \\
\bottomrule
\end{tabular}%
}
\end{table}

\subsection{Out-of-Domain Transfer Setup}
\label{app:transfer}

\paragraph{Tasks.}
\looseness=-1
Database is built from Spider~\citep{yu2018spider}: the assistant receives a database schema and a natural-language request and must produce an SQL query, which is scored against the reference. Relational databases underlie most enterprise data, and learning directly over them is an active direction on its own~\citep{chen2025relgnn}. Actions is built from the Berkeley Function Calling Leaderboard~\citep{2024bfcl}: the assistant receives a tool specification and a user instruction and must emit a programmatic API call, which is checked for an exact abstract-syntax-tree match against the reference. Tool use underlies most LLM agents, and emitting correct API calls from natural-language instructions is an active area of study~\citep{schick2023toolformer}.

\paragraph{Adaptation.}
\looseness=-1
Starting from each model's \ours checkpoint trained on the Math task, we apply a lightweight final-turn-only SFT pass on the target domain. Concretely, for each domain we collect a small multi-turn corpus and supervise only the final-turn answer span. Intermediate-turn behavior is inherited from the math \ours checkpoint without further modification. No \abbr pass is performed.

\begin{table}[t]
\centering
\footnotesize
\setlength{\tabcolsep}{3pt}
\caption{Main results of \Cref{tab:main} with 95\% hierarchical bootstrap confidence intervals over the $103$ test instructions. Accuracy and single-turn attainment $A$ in \%.}
\label{tab:main-ci}
\resizebox{\linewidth}{!}{%
\begin{tabular}{llccccc}
\toprule
Setting & Method & \llamasmall & \llama & \qwen & \olmo & \phimodel \\
\midrule
\multirow{2}{*}{\full}
 & Baseline & $62.5$ [$55.4$, $69.4$] & $72.5$ [$65.9$, $78.7$] & $84.2$ [$79.0$, $88.9$] & $82.9$ [$77.5$, $88.0$] & $88.3$ [$82.7$, $93.2$] \\
 & \ours    & $58.3$ [$51.1$, $65.4$] & $73.8$ [$67.4$, $79.8$] & $83.6$ [$78.2$, $88.5$] & $79.5$ [$73.6$, $85.2$] & $89.8$ [$85.0$, $94.2$] \\
\midrule
\multirow{2}{*}{\concat}
 & Baseline & $60.2$ [$53.1$, $67.2$] & $71.5$ [$64.7$, $78.0$] & $85.0$ [$80.1$, $89.6$] & $82.2$ [$76.4$, $87.6$] & $89.4$ [$84.5$, $93.9$] \\
 & \ours    & $52.6$ [$45.5$, $59.6$] & $72.3$ [$65.8$, $78.5$] & $82.6$ [$77.3$, $87.5$] & $77.8$ [$71.6$, $83.7$] & $88.0$ [$82.7$, $92.6$] \\
\midrule
\multirow{2}{*}{\sharded}
 & Baseline & $39.2$ [$33.0$, $45.8$] & $44.7$ [$38.3$, $51.4$] & $58.9$ [$51.8$, $65.9$] & $54.4$ [$47.6$, $61.1$] & $61.2$ [$54.6$, $67.8$] \\
 & \ours    & $\bm{60.4}$ [$53.6$, $67.2$] & $\bm{71.7}$ [$65.4$, $77.8$] & $\bm{78.7}$ [$73.1$, $84.0$] & $\bm{77.1}$ [$71.4$, $82.5$] & $\bm{82.4}$ [$77.2$, $87.3$] \\
\midrule
\multicolumn{2}{l}{\textbf{Single-turn Attainment} $A$}
 & $100$ [$92$, $109$] & $100$ [$93$, $108$] & $93$ [$87$, $98$] & $94$ [$89$, $99$] & $92$ [$87$, $97$] \\
\bottomrule
\end{tabular}%
}
\end{table}

\section{Confidence Intervals for the Main Results}
\label{app:ci}

\looseness=-1
\Cref{tab:main-ci} repeats \Cref{tab:main} with 95\% confidence intervals from a hierarchical bootstrap: we resample the $103$ test instructions with replacement and, within each resampled instruction, its $10$ simulations. Contrasts and ratios use paired resampling on shared instructions, since all conditions are evaluated on the same instruction set. Three readings follow. First, the \sharded gains are decisively outside noise: the baseline and \ours intervals are disjoint on every backbone, and the lower bound of the paired per-instruction difference is at least $+13.5$ points on all five. Second, both Llama backbones are statistically indistinguishable from full single-turn attainment, whereas the other three backbones measurably fall short, with upper bounds of $97$--$99\%$. The Llama intervals also reach down to $92$--$93\%$, so we do not claim that the two Llama backbones demonstrably differ from the other three. Third, for the single-turn \full and \concat views, the intervals before and after training overlap on every backbone.

\section{Additional Controls}
\label{app:controls}

\paragraph{Non-cooperative User Stress Test.}
\label{app:noncoop}

\looseness=-1
To probe behavior beyond the cooperative setting, we modify the user simulator of~\citet{laban2025llms} so that every user turn containing a value first presents a false value, phrased in varied ways, which the simulated user later retracts and corrects. The simulator is constrained so that the correction always appears after the initial error. \Cref{tab:noncoop-examples} shows two examples. We evaluate the same \ours-trained \llama checkpoint on 20 randomly selected test problems with 10 simulations per problem (200 conversations per condition). Accuracy is $69.5\%$ ($139/200$) under retractions versus $71.0\%$ ($142/200$) on the same problems in the cooperative setting.

\begin{table}[h]
\centering
\footnotesize
\caption{Examples of the retraction-and-correction perturbation applied to value-bearing user turns.}
\label{tab:noncoop-examples}
\begin{tabular}{p{0.3\linewidth}p{0.3\linewidth}p{0.3\linewidth}}
\toprule
Original shard & Perturbed shard (false value) & Later correction \\
\midrule
Lori plans to purchase a pair of shoes worth \$320.00. & Lori plans to purchase a pair of shoes worth \$250.00. & I had the shoe price wrong, they're \$320.00 \\
\addlinespace
Their mother earns double the total of Valerie and her brother's salaries combined. & Their mother earns triple the total of Valerie and her brother's salaries combined. & Sorry, the mother earns double their combined total, not triple \\
\bottomrule
\end{tabular}
\end{table}

\begin{table}[t]
\centering
\footnotesize
\setlength{\tabcolsep}{4pt}

\begin{minipage}[t]{0.50\linewidth}
\centering
\caption{Over-deferral rate (\%) on multi-turn \sharded Math: the fraction of conversations with no answer attempt after all shards have been revealed.}
\label{tab:overdefer}
\vspace{2pt}
\begin{tabular}{lccc}
\toprule
Model       & Baseline          & \ours        & $\Delta$ \\
\midrule
\llamasmall & $17.7$            & $\bm{3.1}$   & $-14.6$ \\
\llama      & $22.2$            & $\bm{3.3}$   & $-18.9$ \\
\qwen       & $\phantom{0}5.0$  & $\bm{0.9}$   & $\phantom{0}{-4.1}$ \\
\olmo       & $12.4$            & $\bm{2.3}$   & $-10.1$ \\
\phimodel   & $13.1$            & $\bm{1.9}$   & $-11.2$ \\
\bottomrule
\end{tabular}
\end{minipage}%
\hfill
\begin{minipage}[t]{0.46\linewidth}
\centering
\caption{Single-turn attainment (\%) restricted to the test instances that the baseline \concat view solves, with 95\% bootstrap confidence intervals.}
\label{tab:teacher-correct}
\vspace{2pt}
\begin{tabular}{lc}
\toprule
Model       & $A$ [95\% CI] \\
\midrule
\llamasmall & $98.1$ [$91.4$, $105.3$] \\
\llama      & $98.2$ [$92.6$, $104.3$] \\
\qwen       & $92.1$ [$87.6$, $\phantom{0}96.6$] \\
\olmo       & $93.0$ [$89.1$, $\phantom{0}97.1$] \\
\phimodel   & $92.1$ [$87.9$, $\phantom{0}96.3$] \\
\bottomrule
\end{tabular}
\end{minipage}

\end{table}

\begin{table}[t]
\centering
\footnotesize
\setlength{\tabcolsep}{4pt}
\caption{Single-turn attainment of \ours against the baseline \concat ceiling ($A$, as in \Cref{tab:main}) and against each model's own post-training \concat ceiling ($A_{\text{post}}$). $^{*}$Above $100\%$ because the \concat accuracy of \llamasmall drops after training.}
\label{tab:apost}
\begin{tabular}{lccccc}
\toprule
 & \llamasmall & \llama & \qwen & \olmo & \phimodel \\
\midrule
$A$ (baseline ceiling)                    & $100\%$       & $100\%$ & $93\%$ & $94\%$ & $92\%$ \\
$A_{\text{post}}$ (post-training ceiling) & $115\%^{*}$   & $99\%$  & $95\%$ & $99\%$ & $94\%$ \\
\bottomrule
\end{tabular}
\end{table}

\begin{table}[t]
\centering
\footnotesize
\setlength{\tabcolsep}{6pt}
\caption{\sharded accuracy (\%) of \llama with two sources of intermediate-turn SFT targets, with 95\% bootstrap confidence intervals. In the self-generated variant no external model writes any supervision target.}
\label{tab:selfgen}
\begin{tabular}{lc}
\toprule
Setting & \sharded accuracy [95\% CI] \\
\midrule
Baseline (no training)                      & $44.7$ [$38.3$, $51.4$] \\
\ours, GPT-4o-mini deferrals (default)      & $71.7$ [$65.4$, $77.8$] \\
\ours, self-generated deferrals             & $70.1$ [$63.9$, $76.1$] \\
\bottomrule
\end{tabular}
\end{table}

\paragraph{Over-deferral.}
\looseness=-1
Under the evaluation protocol, the simulator stops once every shard has been revealed, and a response that is not an answer attempt at that point is scored as incorrect. Over-deferral is therefore penalized directly in accuracy, and the gains in \Cref{tab:main} cannot come at its cost. \Cref{tab:overdefer} quantifies it as the fraction of conversations in which the assistant still produces no answer attempt after all information has been revealed: \ours reduces over-deferral by more than $5\times$ on every backbone, to a residual of $0.9$--$3.3\%$.

\paragraph{Attainment on teacher-solvable instances.}
\looseness=-1
Since the teacher runs on the \concat view, \abbr could in principle propagate the teacher's own errors. \Cref{tab:teacher-correct} recomputes the single-turn attainment after dropping the test instances on which the baseline \concat view is incorrect in all $10$ simulations. Attainment is essentially unchanged from the full-set values in \Cref{tab:main}.

\paragraph{Attainment against the post-training ceiling.}
\looseness=-1
The single-turn attainment in \Cref{tab:main} divides by the baseline's \concat accuracy. \Cref{tab:apost} instead divides by each model's own \concat accuracy after \ours training, denoted $A_{\text{post}}$. \llama stays at $99\%$, so its recovery is clean. \llamasmall exceeds $100\%$ only because its \concat accuracy fell by $7.6$ points after training, which marks a single-turn cost rather than surplus recovery (\Cref{exp:res}).

\paragraph{Self-generated SFT corpus.}
\looseness=-1
To quantify the contribution of external supervision, we rebuild the \llama SFT corpus so that no external model or human-written text serves as a supervision target: all intermediate deferrals are generated by \llama itself instead of GPT-4o-mini, all final-turn targets are the model's own correct \concat rollouts, and the $5\%$ of instructions for SFT that would otherwise fall back to the GSM8K reference solution are excluded (the \abbr corpus is fallback-free by construction). External LLMs are then confined to constructing the sharded user turns, which are verified information-equivalent task inputs and receive no loss. Evaluation is unchanged ($103$ instructions, $10$ simulations each). \Cref{tab:selfgen} shows that after full \ours training the self-generated variant is within $1.6$ points of the default corpus, with overlapping confidence intervals, and remains $25.4$ points above the baseline. The external model is therefore an automation convenience in the data pipeline rather than a hidden teacher of the behavior. What the GPT-4o-mini pipeline adds is intermediate-turn helpfulness, namely requests for the specific missing detail, a conversational-quality property that final accuracy does not measure.

\section{Limitations and Future Work}
\label{app:limitations}

\looseness=-1
We expand on the limitations noted in the main paper. First, the test splits released by~\citet{laban2025llms} contain on the order of $100$ instructions per task, which limits the resolution of recovery-rate estimates. We mitigate this by averaging $n=10$ simulations per instance (amounting to roughly $1{,}000$ scored conversations per model-condition cell), by reporting bootstrap confidence intervals (\Cref{app:ci}), and by reporting consistent trends across five backbones and two out-of-domain tasks, but a larger, more diverse evaluation suite would strengthen the multi-turn-recovery claim. Second, the user simulator is cooperative and non-pivoting: it never changes sub-goals, retracts previously revealed information, or interleaves unrelated content. Our underspecification axis treats all shards as information-monotone (each shard adds, never removes, task-relevant information). Real users sometimes reveal contradictory or partially incorrect details that must be reconciled. Extending \ours to handle these richer conversational dynamics is a natural direction for future work. We view the present results as evidence that the simplest and most prevalent multi-turn failure mode---underspecification---is largely closeable with the model's own single-turn behavior as the only answer supervision. On the method side, \abbr applies its dense distillation signal only at the final turn and leaves intermediate-turn deferral to SFT, which learns it cheaply. Supervising the intermediate turns by self-distillation as well, for instance with the same backbone conditioned on an explicit deferral instruction as the teacher, is a natural extension.

\section{Broader Impact}
\label{app:impact}

\looseness=-1
\ours improves the helpfulness and efficiency of LLMs in everyday underspecified conversations. The deferral behavior it instills---asking for missing information rather than hallucinating it---also has a positive honesty externality: users are less likely to receive confidently wrong answers based on assumed-but-unstated premises. The framework requires only modest compute and can be applied to any open-weight instruction-tuned model. We do not anticipate distinct misuse risks beyond those already present in the base backbones.

\section{System Prompts for Inference-time Interventions}
\label{app:test-time-prompts}

\looseness=-1
The two prompt-based interventions modify only the assistant's system prompt; the user-simulator prompt, strategy classifier, answer extractor, and per-task evaluation prompts are all inherited unchanged from~\citet{laban2025llms}. The two judge-gated interventions (\gatedself and \gatedexternal) likewise reuse the \lic default assistant system prompt for the answer pass; the auxiliary judge pass uses a short rubric prompt that asks whether the conversation so far contains sufficient information to attempt a final answer, returning \texttt{YES} or \texttt{NO}. The exact system prompts used for \promptrule and \promptselfcheck on the math task are reproduced below.

\paragraph{\promptrule.}
\begin{verbatim}
You are solving a mathematical problem where the user will provide
information gradually across multiple messages.

CRITICAL RULES:
1. Do NOT assume, fabricate, or guess any values not explicitly stated
   by the user. If a quantity is unknown, state that it is unknown --
   never substitute a hypothetical value or use phrases like "let's
   assume" or "for example, $200 each".
2. Do NOT compute a final numerical answer until ALL quantities required
   for the calculation have been explicitly provided by the user. If
   any value is still missing, withhold computation and instead state
   what information is still needed.
3. When information is incomplete, respond with "I still need the
   following information to solve this: [list]" rather than filling in
   gaps yourself.
4. HOWEVER, as soon as you have enough explicitly stated values to
   perform the calculation, you MUST compute and present the final
   numerical answer immediately. Do not hesitate or ask for
   confirmation -- if the math can be done with the given values, do it.
\end{verbatim}

\paragraph{\promptselfcheck.}
\begin{verbatim}
You are solving a mathematical problem where the user will provide
information gradually across multiple messages.

Before EACH response, you must first determine if you have ALL the
numerical values and information needed to compute the final answer.

Start every response with exactly one of:
READY: NO
READY: YES

If READY: NO -- State that information is still missing. Do NOT attempt
any calculations and do NOT assume or fabricate any missing values.

If READY: YES -- Immediately compute the final answer using ONLY values
explicitly stated by the user. Show your work and end with
FINAL ANSWER: <number>
\end{verbatim}

\end{document}